\documentclass{article}

    \PassOptionsToPackage{numbers, compress}{natbib}
\usepackage[preprint]{neurips_2026}

\usepackage[utf8]{inputenc} 
\usepackage[T1]{fontenc}    
\usepackage{hyperref}       
\usepackage{url}            
\usepackage{booktabs}       
\usepackage{amsfonts}       
\usepackage{nicefrac}       
\usepackage{microtype}      
\usepackage{xcolor}         

\def\eqref#1{Eq.~(\ref{#1})}
\usepackage{thmtools,}
\usepackage{amsmath,amssymb,amsthm}
\usepackage{mathtools}
\usepackage{graphicx}       
\usepackage{booktabs}
\usepackage{algorithm}
\usepackage{algpseudocode}
\usepackage{multirow}
\usepackage{float}
\usepackage{caption}
\usepackage{wrapfig}
\usepackage{pifont}
\usepackage{color, colortbl}

\newcommand{\vx}{\mathbf{x}}
\newcommand{\vy}{\mathbf{y}}
\newcommand{\vz}{\mathbf{z}}

\newcommand{\vv}{\mathbf{v}}

\newcommand{\vn}{\mathbf{n}}

\newcommand{\epsilonb}{{\boldsymbol{\epsilon}}}
\newcommand{\gA}{\mathcal{A}}
\newcommand{\gN}{\mathcal{N}}
\newcommand{\gD}{\mathcal{D}}

\def\eqref#1{Eq.(\ref{#1})}

\newcommand{\cmark}{\ding{51}} 
\newcommand{\xmark}{\ding{55}} 

\definecolor{LightBlue}{rgb}{0.85, 0.90, 0.94}
\newcommand{\cg}{\cellcolor{LightBlue}}

\title{FlowLPS: Langevin-Proximal Sampling for Flow-based Inverse Problem Solvers}

\author{%
  Jonghyun Park \\
  KAIST \\
  \texttt{jhpark99@kaist.ac.kr} \\
  \And
  Jong Chul Ye \\
  KAIST \\
  \texttt{jong.ye@kaist.ac.kr} \\
}

\begin{document}

\maketitle

\begin{figure}[h]
    \centering
    \includegraphics[width=\linewidth]{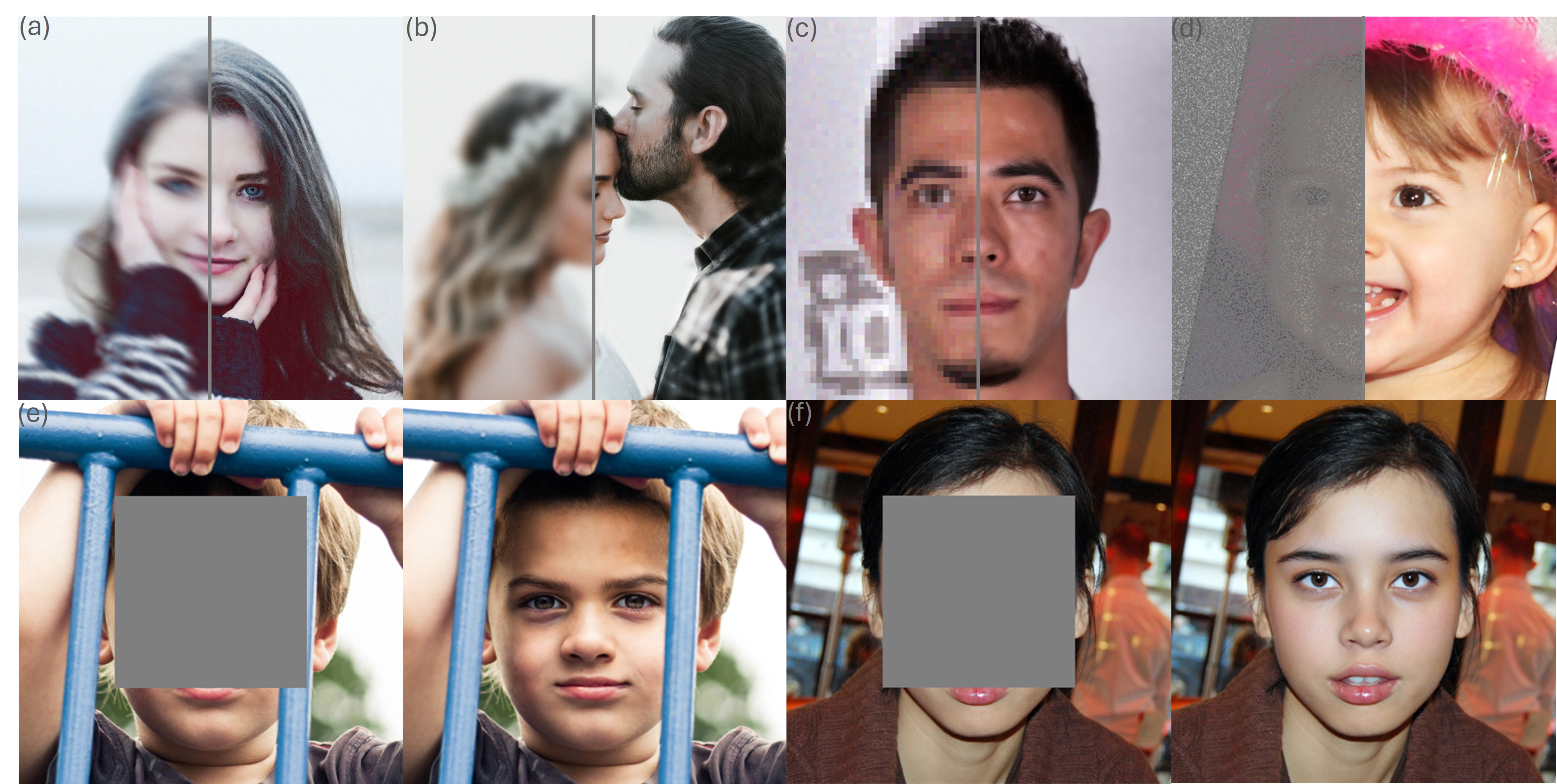}
    \caption{\textbf{Qualitative results of FlowLPS}. (a) Motion deblurring, (b) Gaussian deblurring, (c) Super-resolution ($\times12$) and (d) random inpainting results before and after FlowLPS. (e) and (f) are results for box inpainting.} 
     \label{fig:main}
     \vspace{-0.2cm}
\end{figure}

\begin{abstract}
Deep generative models are powerful priors for imaging inverse problems, but training-free solvers for latent flow models face a practical finite-step trade-off.
Optimization-heavy methods quickly improve measurement consistency, but in highly nonlinear latent spaces, their results can depend strongly on where local refinement is initialized, often degrading perceptual realism.
In contrast, stochastic sampling methods better preserve posterior exploration, but often require many iterations to obtain sharp, measurement-consistent reconstructions.
To address this trade-off, we propose \textit{FlowLPS}, a training-free latent flow inverse solver based on Langevin-Proximal Sampling.
At each reverse step, FlowLPS uses a few Langevin updates to perturb the model-predicted clean estimate in posterior-oriented directions, providing stochastic initializations for local refinement.
It then applies local MAP-style proximal refinement to rapidly improve measurement consistency from the Langevin-updated estimate.
We additionally use controlled pCN-style re-noising to stabilize the reverse trajectory while retaining trajectory coherence.
Experiments on FFHQ and DIV2K across five linear inverse problems show that FlowLPS achieves a strong balance between measurement fidelity and perceptual quality, with additional experiments on pixel-space inverse problems and phase retrieval.
\end{abstract}

\section{Introduction}

The goal of solving inverse problems is to recover underlying signals $\vx_0$ from noisy measurements $\vy$:
\begin{align}
    \vy = \gA(\vx_0) + \vn
    \label{eq:forward}
\end{align}
where $\gA$ is the forward measurement operator and $\vn \sim \gN(\mathbf{0}, \sigma_n^2\mathbf{I})$ denotes measurement noise. 
Due to the ill-posed nature of inverse problems, multiple solutions may satisfy \eqref{eq:forward}. Hence, the objective is to find a meaningful solution that aligns with both the true data distribution and the measurement.

Recent diffusion and flow models~\citep{ho2020denoising, song2020Score, lipman2023flow, liu2023flow, esser2024scaling} have become powerful pretrained priors for inverse problems.
Among generative inverse solvers, two prominent lines of work have developed: 
posterior-guided inference and optimization-based correction.
Posterior-guided methods approximate measurement-conditioned posteriors~\citep{chung2023diffusion, song2023pseudoinverseguided, he2024manifold, rout2024solving, chung2024decomposed, zhang2024improving, pokle2024otode, patel2024steering, kim2025flowdps}, while optimization-based methods enforce measurement consistency through variational objectives, direct optimization, or proximal corrections~\citep{mardani2023variational, zhu2023denoising, zilberstein2024repulsive, martin2025pnpflow, erbach2025flair, li2025decoupled, pourya2025flower}.

However, in latent-space inverse solving under a finite reverse-time budget, the practical behavior of a solver is often determined not only by its nominal category, but by how it allocates computation between stochastic posterior exploration and measurement-driven refinement.
This issue becomes especially pronounced in latent generative models such as Stable Diffusion, where nonlinearity of the VAE decoder can make a single measurement correction at each reverse step insufficient~\citep{rout2024solving, song2024solving}.
Several recent methods therefore strengthen measurement enforcement, including hard data-consistency optimization in ReSample~\citep{song2024solving}, decomposition-based correction in FlowDPS~\citep{kim2025flowdps}, and decoupled exact-consistency optimization in FLAIR~\citep{erbach2025flair}.

From this practical finite-step perspective, latent-space solvers exhibit two representative operating regimes.
\emph{Refinement-dominant} solvers emphasize repeated local correction.
These methods~\citep{chung2024decomposed, erbach2025flair, pourya2025flower} can rapidly reduce measurement mismatch, but in nonlinear latent spaces, their refinement behavior can become sensitive to the initialization and may perturb the prior-supported trajectory, leading to artifacts or degraded realism~\citep{chung2022improving, zirvi2025diffusionstate}.
Conversely, \emph{sampling-dominant} solvers, such as Langevin-based posterior samplers~\citep{zhang2024improving}, preserve stochastic posterior exploration and prior fidelity, but often require many iterations before concentrating into sharp, measurement-consistent estimates.
Under practical reverse-time budgets, these two extremes expose a central trade-off between stochastic exploration and local measurement-consistent refinement.

In this work, we propose \textbf{FlowLPS} (Langevin-Proximal Sampling), a training-free hybrid solver for latent flow-based inverse problems.
FlowLPS targets the practical finite-step gap between refinement-dominant and sampling-dominant solvers.
Motivated by the Langevin+correction viewpoint, FlowLPS adapts stochastic exploration followed by deterministic correction to latent flow inference.
The key idea is to use Langevin updates not as a long-run posterior sampler, but as a stochastic initialization mechanism for local refinement.
At each reverse step, FlowLPS applies a few Langevin updates to perturb the model-predicted clean estimate in posterior-oriented directions, followed by local MAP-style proximal refinement to improve measurement consistency from this initialization.
This avoids relying solely on deterministic refinement from a fixed initialization, while also avoiding the high cost of extensive stochastic sampling before obtaining a sharp measurement-consistent estimate.

To stabilize the reverse trajectory after refinement, we further use controlled pCN-style re-noising~\citep{beskos2017geometric} to refresh the noise component while retaining useful trajectory coherence.
Together, these components provide a practical exploration--refinement balance for latent inverse solving under a fixed computational budget.

Our key contributions are:
\begin{itemize}
    \item We propose \textbf{FlowLPS}, a training-free latent flow inverse solver that combines posterior-oriented Langevin initialization with local measurement-consistent proximal refinement in a practical finite-step inference procedure.

    \item We provide a local analysis explaining the complementary roles of Langevin exploration and proximal refinement, validated through ablations on the number of Langevin steps.
    
    \item We introduce controlled pCN-style re-noising that stabilizes the reverse trajectory by injecting stochasticity while retaining trajectory coherence.
    
    \item We show that FlowLPS achieves a strong balance between measurement fidelity and perceptual quality on FFHQ and DIV2K, and further validate its applicability through additional experiments on pixel-space inverse problems and phase retrieval.
\end{itemize}

\begin{figure}[t]
    \centering
    \includegraphics[width=0.95\linewidth]{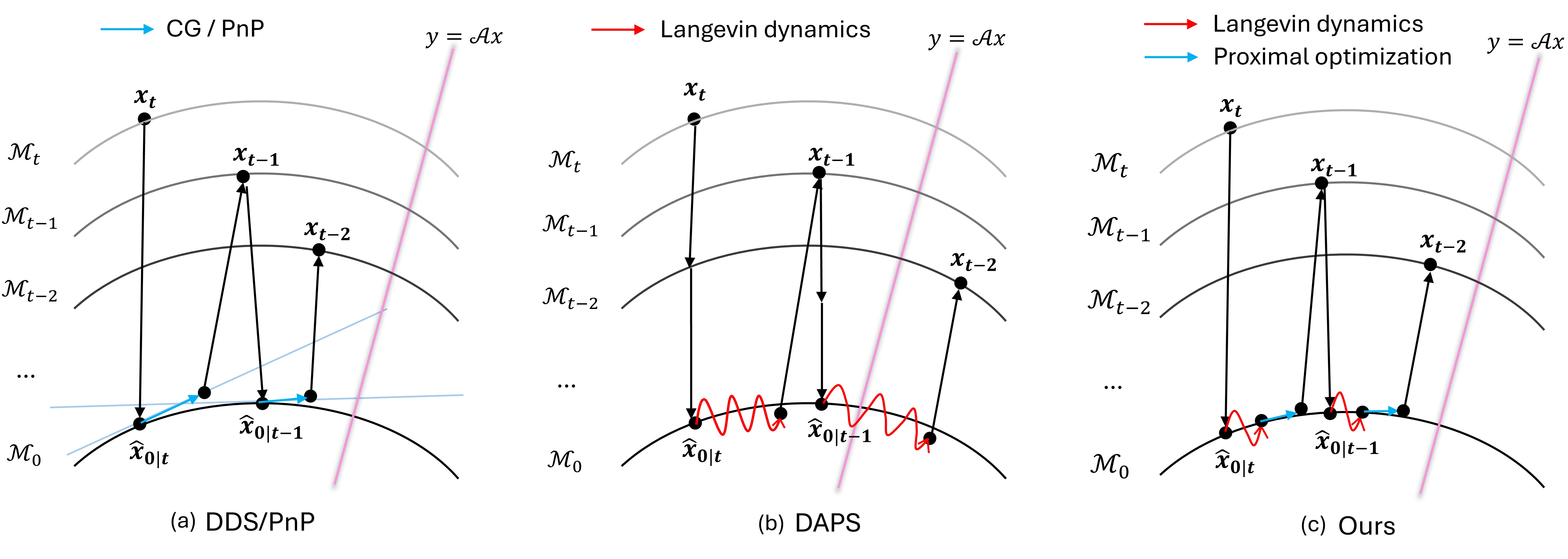}
    \vspace{-0.5em}
    \caption{
\textbf{Practical finite-step operating regimes for inverse solvers.}
(a) Refinement-dominant solvers, such as DDS/PnP, repeatedly apply measurement correction, which can be initialization-dependent and may induce large local deviations in nonlinear latent spaces.
(b) Sampling-dominant solvers, such as DAPS, rely on Langevin exploration, preserving stochasticity but often requiring many iterations to concentrate near measurement-consistent estimates.
(c) FlowLPS combines both: Langevin updates provide stochastic posterior-oriented initialization, and proximal refinement applies a shorter local correction toward measurement consistency.
}
    \vspace{-0.5em}
    \label{fig:structure}
\end{figure}


\section{Related Work}
\paragraph{Diffusion and Flow-Based Inverse Solvers.}
Generative inverse solvers are often discussed through two broad lines: posterior-guided inference and optimization-based correction.
Posterior-guided methods steer the generative trajectory using approximations to 
$\nabla_{\vz_t} \log p(\vz_t|\vy) = \nabla_{\vz_t}\log p(\vz_t) + \nabla_{\vz_t}\log p(\vy|\vz_t)$,
where $\vz_t$ denotes an intermediate state.
Since the likelihood term is generally intractable, DPS~\citep{chung2023diffusion} and related methods~\citep{song2023pseudoinverseguided, rout2024solving, pokle2024otode} approximate this term using denoised estimates or Gaussian assumptions on $p(\vz_0|\vz_t)$.
Later methods reduce direct guidance cost by updating the denoised estimate $\hat{\vz}_{0|t}$, which has become a practical surrogate for posterior guidance~\citep{he2024manifold, chung2024decomposed, zhang2024improving,patel2024steering, kim2025flowdps}.
Optimization-based methods enforce measurement consistency through explicit optimization, variational objectives, or proximal corrections~\citep{mardani2023variational, zhu2023denoising, zilberstein2024repulsive,  martin2025pnpflow, erbach2025flair, li2025decoupled, pourya2025flower}.
These methods can rapidly reduce measurement error, especially with multiple refinement steps per reverse step.
However, in latent-space inverse problems, decoder-induced nonlinearity makes the refinement landscape highly nonconvex, so aggressive local correction can be initialization-sensitive and may degrade perceptual realism.

\paragraph{Operating Regimes in Latent-Space Solvers.}
Beyond this methodological distinction, we focus on the practical finite-step behavior of latent-space inverse solvers.
From this operational viewpoint, methods often exhibit two representative regimes.

\emph{Refinement-dominant solvers} prioritize measurement consistency through repeated gradient refinement, Krylov/CG-style correction, or proximal optimization.
For example, DDS~\citep{chung2024decomposed} performs multi-step refinement, while FLAIR~\citep{erbach2025flair} and Flower~\citep{pourya2025flower} use strong consistency or proximal updates.
Such methods can quickly reduce measurement mismatch, but deterministic refinement may be initialization-dependent and can perturb the prior-supported trajectory in nonlinear latent spaces.
\emph{Sampling-dominant solvers} prioritize stochastic posterior exploration.
A representative example is DAPS~\citep{zhang2024improving}, which uses Langevin dynamics to sample from $p(\vz_0|\vz_t,\vy)$.
This regime better preserves stochasticity and prior fidelity, but often requires many iterations to concentrate samples into sharp, measurement-consistent estimates.
Under matched budgets, this can limit reconstruction fidelity.

FlowLPS is motivated by this practical finite-step tension. 
It uses a few Langevin updates to obtain posterior-oriented initializations, followed by local proximal refinement to improve measurement consistency.
This places FlowLPS between the two regimes as an exploration--refinement hybrid.


\section{Preliminaries}

\paragraph{Flow Matching.}
Flow models generate samples by transporting a source distribution $p_1=p$ to a target distribution $p_0=q$ through a time-dependent path $p_t$ governed by the reverse-time ODE
\begin{align}
    \label{eq:flow_ode}
    d\vz_t = \vv_t(\vz_t)dt, \qquad \vz_1 \sim p_1=p,
\end{align}
where $\vv_t:[0,1]\times \mathbb{R}^d \to \mathbb{R}^d$ is the marginal vector field.

For rectified flow~\citep{liu2023flow}, the interpolation is given by
\begin{equation}
    \vz_t = t\vz_1 + (1-t)\vz_0,
\end{equation}
where $(\vz_0,\vz_1)$ is drawn from the independent coupling $\pi(\vz_0,\vz_1)=q(\vz_0)p(\vz_1)$.
The corresponding marginal vector field satisfies
\begin{align}
    \label{eq:vector_field}
    \vv_t(\vz_t) = \mathbb{E}[\vz_1-\vz_0|\vz_t]
    = \mathbb{E}[\vz_1|\vz_t] - \mathbb{E}[\vz_0|\vz_t].
\end{align}
Therefore, the posterior mean of the clean image $\mathbb{E}[\vz_0|\vz_t]$ and the noise $\mathbb{E}[\vz_1|\vz_t]$ can be written as
\begin{align}
    \hat\vz_{0|t}:=\mathbb{E}[\vz_0|\vz_t] = \vz_t - t\vv_t(\vz_t), & \quad  \hat\vz_{1|t}:=\mathbb{E}[\vz_1|\vz_t] = \vz_t +(1-t) \vv_t(\vz_t).
\end{align}
Accordingly, the Euler discretization of the ODE~\eqref{eq:flow_ode} can be written as
\begin{align}
    \vz_{t-\Delta t} &= \vz_{t} -\Delta t \left( \mathbb{E}[\vz_1|\vz_t]  - \mathbb{E}[\vz_0|\vz_t]\right) \notag \\
    &= \big(1-(t-\Delta t)\big) \mathbb{E}[\vz_0|\vz_t]+(t-\Delta t)\mathbb{E}[\vz_1|\vz_t]. \label{eq:update}
\end{align}

\paragraph{Langevin+correction viewpoint.}
Many posterior samplers combine stochastic Langevin exploration with a deterministic correction step.
A useful abstraction is
\begin{align}
    \tilde{\vz}^{k+1} &= \vz^k - \gamma \nabla f(\vz^k) + \sqrt{2\gamma}\,\epsilon^k,
    \qquad \epsilon^k \sim \mathcal{N}(\mathbf{0},\mathbf{I}), \\
    \vz^{k+1} &= \mathcal{C}_\gamma(\tilde{\vz}^{k+1}).
\end{align}
Different choices of $\mathcal{C}_\gamma$ recover different methods:
$\mathcal{C}_\gamma=\mathrm{Id}$ gives ULA,
$\mathcal{C}_\gamma=\Pi_{\mathcal{C}}$ gives projected Langevin algorithms~\citep{bubeck2018sampling, lamperski2021projected}, and $\mathcal{C}_\gamma=\mathrm{prox}_{\gamma g}$ gives proximal-Langevin-type updates~\citep{salim2020primal, ehrhardt2024proximal, renaud2025stability}.

FlowLPS is inspired by this Langevin+correction viewpoint, but differs in goal and setting from classical proximal Langevin samplers.
While classical methods study posterior convergence under specific assumptions, FlowLPS uses the same operational principle as a practical finite-step mechanism for latent inverse solving:
Langevin updates provide stochastic posterior-oriented initialization, and proximal refinement provides local measurement-consistent correction.


\section{Method}
\label{sec:method}

We present FlowLPS as a practical finite-step latent flow solver for inverse problems.
Motivated by the Langevin+correction viewpoint, FlowLPS adapts an exploration--refinement structure to latent flow inference.

At each reverse-time step, FlowLPS performs three operations.
First, a few Langevin updates produce a stochastic, posterior-oriented clean image estimate.
Second, a local MAP-style proximal refinement improves measurement consistency from this stochastic initialization.
Third, a pCN-style re-noising refreshes the noise component in a controlled manner, and the affine flow update advances the trajectory.
Together, these components combine stochastic posterior-oriented initialization with local measurement-consistent refinement.

\subsection{Phase 1: Posterior-Oriented Langevin Initialization}
\label{sec:langevin}

At timestep $t$, we adopt a local Gaussian approximation of the conditional clean distribution, following prior work~\citep{song2023pseudoinverseguided, zhang2024improving, pokle2024otode}:
\begin{equation}
    p(\vz_0|\vz_t) \approx \mathcal{N}(\hat\vz_{0|t}, s_t^2\mathbf{I}),
\end{equation}
where $s_t^2$ is specified heuristically.
Under this approximation, the resulting conditional posterior gradient can be evaluated as
\begin{equation}
    \nabla_{\vz_0}\log p(\vz_0|\vz_t,\vy)
    =\nabla_{\vz_0} \log p(\vy|\vz_0) + \nabla_{\vz_0}\log p(\vz_0|\vz_t).
\end{equation}
Starting from the clean image estimate $\hat{\vz}_{0|t}^{(0)}=\hat\vz_{0|t}$, we run $N_L$ Langevin updates:
\begin{align}
    \hat{\vz}_{0|t}^{(j+1)} = \hat{\vz}_{0|t}^{(j)} 
    + \eta_t\nabla_{\hat{\vz}_{0|t}^{(j)}} \log p(\hat{\vz}_{0|t}^{(j)}|\vz_t,\vy)+\sqrt{2\eta_t}\,\epsilonb_j,
    \quad \epsilonb_j\sim\mathcal{N}(\mathbf{0},\mathbf{I}),
    \label{eq:flowlps_langevin}
\end{align}
for $j=0,\cdots,N_L-1$.
This step injects stochastic posterior-oriented exploration around the model-induced trajectory, producing diverse, posterior-informed initializations for the subsequent deterministic refinement.

\subsection{Phase 2: Local MAP-Style Proximal Refinement}
\label{sec:proximal}

Finite-step Langevin updates provide posterior-oriented exploration, but may not yield a measurement-consistent estimate within a limited number of steps.
We therefore refine the Langevin-updated point via a local MAP problem.

\begin{restatable}{prop}{mapprox}
    \label{prop:mapprox}
    Assume $p(\vz_0|\vz_t)=\mathcal{N}(\hat{\vz}_{0|t}, s_t^2 I)$ and Gaussian measurement noise $\vn\sim\mathcal{N}(\mathbf{0},\sigma_n^2 \mathbf{I})$.
    Then, the maximizer is given by
    \begin{align}
        \label{eq:proximal_objective}
        \vz_{0\mid t}^\ast
        = \arg\max_{\vz_0}\log p(\vz_0|\vz_t,\vy)
        = \arg\min_{\vz_0}\|\vy-\mathcal{A}(\vz_0)\|^2
        +\frac{\sigma_n^2}{s_t^2}\|\vz_0-\hat{\vz}_{0|t}\|^2 .
    \end{align}
\end{restatable}

Proposition~\ref{prop:mapprox} reformulates conditional MAP estimation as proximal measurement refinement.
In FlowLPS, we optimize this objective for a finite number of gradient steps initialized at the Langevin-updated point $\hat{\vz}_{0|t}^{(N_L)}$.
This stochastic initialization can change the attraction region from which refinement is performed, while the proximal objective improves measurement consistency locally.

FlowLPS therefore interpolates between two regimes: removing the proximal step ($N_P=0$) yields a sampling-dominant solver similar to DAPS~\citep{zhang2024improving}, while removing the Langevin step ($N_L=0$) yields a refinement-dominant solver similar to Flower~\citep{pourya2025flower}.

\subsection{Local Finite-Step Analysis}
\label{sec:local_analysis}

We provide a local, interpretive analysis of the exploration--refinement mechanism.
This analysis focuses on within-basin concentration after the Langevin phase provides a useful local initialization, rather than on global basin selection.
Specifically, we consider a neighborhood where the negative log conditional posterior is well approximated by a smooth strongly convex objective.
Under this local basin model, deterministic refinement contracts toward the local minimizer, whereas continued Langevin updates retain an additive stochastic variance term.
This explains why proximal refinement can concentrate faster than continuing stochastic Langevin updates once a useful initialization is obtained, and motivates the $N_L$ ablations in Sec.~\ref{langevin_effect}.

Define $\Phi(\vz_0):=-\log p(\vz_0|\vz_t,\vy)$. 
Under the local Gaussian approximation in Sec.~\ref{sec:langevin}, $\Phi$ is equivalent to the proximal objective in \eqref{eq:proximal_objective} up to additive constants and a positive scaling factor. 
Thus, both deterministic refinement and ULA can be analyzed locally through the same smooth objective $\Phi$ around a local minimizer $\vz_B^\star$.

\begin{restatable}{prop}{captureexpansion}
\label{prop:capture_expansion}
Let $\Phi$ be twice differentiable on a convex neighborhood $\Omega_B$ of a local minimizer
$\vz_B^\star$, and assume
\begin{equation}
    \mu \mathbf{I}\preceq \nabla^2\Phi(\vz)\preceq L\mathbf{I}, \qquad \forall \vz\in\Omega_B,
\end{equation}
for some $0<\mu\le L$.
Let $R_N$ denote $N$ gradient-refinement steps with step size $0<\tau\le \frac{1}{L}$,
and assume the iterates remain in $\Omega_B$.
Then, with $q=1-\tau\mu<1$,
\begin{equation}
    \|R_N(\vz)-\vz_B^\star\|
    \le
    q^N\|\vz-\vz_B^\star\|.
\end{equation}
\end{restatable}

Consequently, for any $B_r=\{\vz:\|\vz-\vz_B^\star\|\le r\}$ with $B_{r/q^N}\subset\Omega_B$, we have
\begin{equation}
    B_{r/q^N}\subseteq R_N^{-1}(B_r).
\end{equation}
Thus, refinement enlarges the effective pre-refinement region from which a small neighborhood of the local minimizer can be reached.

\begin{restatable}{lemma}{ulanoisefloor}
\label{lem:ula_noise_floor}
Under the assumptions of Proposition~\ref{prop:capture_expansion}, consider the ULA
update
\begin{equation}
    \vz_{k+1}
    =\vz_k-\tau\nabla\Phi(\vz_k)+\sqrt{2\tau}\boldsymbol{\xi}_k,
    \quad \boldsymbol{\xi}_k\sim\mathcal{N}(\mathbf{0},\mathbf{I}_d),
\end{equation}
and assume the iterates remain in $\Omega_B$. Then
\begin{equation}
    \mathbb{E}\|\vz_N-\vz_B^\star\|^2
    \le
    q^{2N}\|\vz_0-\vz_B^\star\|^2
    +
    2\tau d\,\frac{1-q^{2N}}{1-q^2},
    \qquad q=1-\tau\mu.
\end{equation}
\end{restatable}

Together, Proposition~\ref{prop:capture_expansion} and Lemma~\ref{lem:ula_noise_floor}
clarify the complementary local behavior of deterministic refinement and Langevin exploration under this local basin model.
Deterministic refinement contracts geometrically toward the local minimizer and enlarges the effective pre-refinement region by a factor $q^{-N}$.
In contrast, ULA has the same contraction in its drift, but the injected Gaussian noise introduces an additive variance term in the finite-step bound.
Thus, Langevin updates maintain stochastic exploration but can remain diffuse over a limited number of steps, whereas proximal refinement rapidly concentrates once initialized in a useful local region.

This local view suggests a qualitative update-allocation trade-off controlled by $N_L$.
Too few Langevin updates make refinement strongly tied to the model-predicted initialization, while too many Langevin updates leave insufficient update budget for local measurement-consistent refinement.
We evaluate this trade-off through the $N_L$ ablations in Sec.~\ref{langevin_effect} and Appendix~\ref{sec:N_L_ablation}.

\vspace{-0.05cm}
\subsection{Controlled pCN-Style Re-Noising}
\label{sec:pcn_renoising}
\vspace{-0.05cm}
After obtaining the refined clean estimate $\vz_{0|t}^{\ast}$, a standard way to move to the next noise level is
\begin{equation}
    \vz_{t-\Delta t} = (1-(t-\Delta t)) \vz_{0|t}^{\ast}+(t-\Delta t)\epsilonb, \quad \epsilonb\sim \mathcal{N}(\mathbf{0}, \mathbf{I}).
\end{equation}
Rather than drawing this noise component at each step, FlowLPS uses a pCN-style refresh and substitutes the refreshed noise into the re-noising equation.
Given $\hat{\vz}_{1|t}^{(i)}\sim\mathcal{N}(\mathbf{0},\mathbf{I})$, the pCN update is
\begin{align}
    \hat{\vz}_{1|t}^{(i+1)}
    = \rho_t \hat{\vz}_{1|t}^{(i)} + \sqrt{1-\rho_t^2}\,\epsilonb, \quad \epsilonb\sim \mathcal{N}(\mathbf{0}, \mathbf{I}).
    \label{eq:pCN_practical}
\end{align}
This form is inspired by the Gaussian-preserving property of pCN~\citep{beskos2017geometric}.
In practice, FlowLPS initializes the refresh from the model-predicted noise
$\hat{\vz}_{1|t}^{(0)}=\vz_t+(1-t)\vv_t(\vz_t)$, trading exact marginal preservation for trajectory coherence.
We use a single pCN-style re-noising step with mixing coefficient $\rho_t=\sqrt{1-t}$, which adaptively balances fresh stochasticity with preservation of the model-predicted noise direction. Ablations on the mixing schedule and number of pCN steps are provided in Sec.~\ref{sec:rho_t strategy} and Appendix~\ref{num_of_pCN_steps}.

\vspace{-0.1cm}
\paragraph{Diversity.}
Although proximal refinement locally concentrates each trajectory, FlowLPS remains stochastic through both Langevin updates and pCN-style re-noising.
We evaluate the resulting diversity--validity trade-off in Appendix~\ref{sec:sample_diversity}.

The complete FlowLPS algorithm is summarized in Algorithm~\ref{alg:FlowLPS}.

\begin{algorithm}[t]
\small
\caption{Algorithm of FlowLPS}\label{alg:FlowLPS}
    \begin{algorithmic}
        \State {\bfseries Require:} $\vy$, $\mathcal{A}$, VAE decoder $\mathcal{D}$, $\vv_t^\theta$, $N_L$
        \State {\bfseries Initialize:} $\vz_1 \sim \mathcal{N}(\mathbf{0},\mathbf{I})$
        \For{$t: 1 \rightarrow 0$}
            \State $\vv_t \leftarrow \vv_t^\theta(\vz_t)$ 
            \State $\hat{\vz}_{0|t}^{(0)}\leftarrow \vz_t -t\vv_t, \; \hat{\vz}_{1|t}\leftarrow \vz_t+(1-t)\vv_t$ 
            \State $\hat\epsilonb_{1|t} \leftarrow \sqrt{1-(t-\Delta t)}\hat{\vz}_{1|t}+\sqrt{t-\Delta t}\epsilonb$, \;\;$\epsilonb\sim \mathcal{N}(\mathbf{0},\mathbf{I})$ \Comment{pCN update}
            \For{$j=0, 1, \cdots, N_L-1$}
                \vspace{0.5pt}
                \State $\hat{\vz}_{0|t}^{(j+1)}\leftarrow \hat{\vz}_{0|t}^{(j)}+\eta_t\nabla_{\hat\vz_{0|t}^{(j)}} \log p(\hat{\vz}_{0|t}^{(j)}|\vz_t,\vy)+\sqrt{2\eta_t}Z_j, \quad Z_j\sim \mathcal{N}(\mathbf{0},\mathbf{I})$ \Comment{Langevin dynamics}
            \EndFor 
            \State $\vz \leftarrow \hat{\vz}_{0|t}^{(N_L)}$
            \State $\vz_{0|t}^* \leftarrow \underset{\vz}{\text{argmin}}\; ||\vy-\gA(\gD(\vz))||^2+\frac{\sigma_n^2}{s_t^2}||\vz-\hat{\vz}_{0|t}^{(N_L)}||^2$  \Comment{Proximal optimization}
            \State $\vz_{t-\Delta t}\leftarrow \big(1-(t-\Delta t)\big)\vz_{0|t}^*+(t-\Delta t)\hat\epsilonb_{1|t}$
        \EndFor
        \State \Return $\mathcal{D}(\vz_{0|t}^{\ast})$
    \end{algorithmic}
\end{algorithm}

\begin{table*}[t]
  \caption{\textbf{Quantitative Results on FFHQ 1k and DIV2K 0.8k datasets} with $\sigma_n=0.03$. \textbf{Bold}: best. \underline{Underline}: second best.}
  \scriptsize
  \label{whole results}
  \centering
  \resizebox{\linewidth}{!}{
  \begin{tabular}{l|l|cccc|cccc}
    \toprule
    \multirow{2}{*}{\textbf{Task}} & \multirow{2}{*}{\textbf{Method}} & \multicolumn{4}{c|}{\textbf{FFHQ 1k}} & \multicolumn{4}{c}{\textbf{DIV2K 0.8k}} \\
    & & \textbf{PSNR} ($\uparrow$) & \textbf{SSIM} ($\uparrow$) & \textbf{FID} ($\downarrow$) & \textbf{LPIPS} ($\downarrow$) & \textbf{PSNR} ($\uparrow$) & \textbf{SSIM} ($\uparrow$) & \textbf{FID} ($\downarrow$) & \textbf{LPIPS} ($\downarrow$) \\
    \midrule
    \multirow{7}{*}{Inpainting (Box)}
    & DAPS & \underline{21.78} & 0.842 & 24.75 & \underline{0.114} & 23.72 & 0.841 & \underline{19.96} & 0.080 \\ 
    & RLSD & 16.32 & 0.519 & 191.6 & 0.538 & 16.25 & 0.417 & 164.1 & 0.535 \\
    & FlowDPS & 18.81 & 0.771 & 58.49 & 0.204 & 21.36 & 0.756 & 36.13 & 0.142 \\
    & FlowChef & 17.72 & 0.701 & 74.07 & 0.265 & 21.02 & 0.688 & 59.44 & 0.160 \\
    & FLAIR & \textbf{21.84} & 0.836 & \underline{22.63} & 0.122 & \underline{23.90} & 0.845 & 20.76 & \underline{0.076} \\
    & Flower & 21.63 & \textbf{0.876} & 48.91 & 0.119 & \textbf{24.41} & \textbf{0.868} & 22.38 & 0.081 \\
    & \cg Ours & \cg 21.68 & \cg \underline{0.849} & \cg \textbf{21.90} & \cg \textbf{0.107} & \cg 23.73 & \cg \underline{0.862} & \cg \textbf{15.50} & \cg \textbf{0.065} \\
    \midrule
    \multirow{7}{*}{Inpainting (Random)}
    & DAPS & \underline{33.41} & \underline{0.857} & \underline{13.28} & \underline{0.034} & 27.97 & 0.799 & 15.37 & 0.039 \\ 
    & RLSD & 17.04 & 0.539 & 123.6 & 0.441 & 15.63 & 0.392 & 137.7 & 0.480 \\
    & FlowDPS & 30.36 & 0.808 & 32.03 & 0.109 & 25.44 & 0.713 & 27.96 & 0.114 \\
    & FlowChef & 32.07 & 0.849 & 63.23 & 0.077 & 25.35 & 0.688 & 57.82 & 0.139 \\
    & FLAIR & 14.09 & 0.502 & 131.9 & 0.443 & 13.78 & 0.355 & 116.9 & 0.408 \\
    & Flower & 31.37 & 0.826 & 15.76 & 0.048 & \underline{28.81} & \textbf{0.843} & \underline{9.824} & \underline{0.026} \\
    & \cg Ours & \cg \textbf{34.34} & \cg \textbf{0.886} & \cg \textbf{7.460} & \cg \textbf{0.019} & \cg \textbf{28.85} & \cg \underline{0.841} & \cg \textbf{9.360} & \cg \textbf{0.025} \\
    \midrule
    \multirow{7}{*}{Gaussian Deblurring}
    & DAPS & 27.21 & 0.715 & 41.72 & 0.171 & 21.42 & 0.498 & 81.81 & 0.294 \\ 
    & RLSD & 21.42 & 0.626 & 110.3 & 0.426 & 17.90 & 0.424 & 154.9 & 0.571 \\
    & FlowDPS & 26.87 & 0.715 & 45.19 & 0.201 & 21.17 & 0.519 & 78.79 & 0.322 \\
    & FlowChef & 24.87 & 0.690 & 46.61 & 0.260 & 21.19 & 0.542 & 102.6 & 0.363 \\
    & FLAIR & 26.84 & 0.704 & \textbf{29.03} & \underline{0.165} & 21.27 & 0.509 & \underline{63.97} & \underline{0.276} \\
    & Flower & \textbf{28.67} & \textbf{0.796} & 94.34 & 0.167 & \underline{21.84} & \textbf{0.561} & 88.65 & 0.312 \\
    & \cg Ours & \cg \underline{27.76} & \cg 0.737 & \cg \underline{31.22} & \cg \textbf{0.130} & \cg \textbf{21.97} & \cg \underline{0.549} & \cg \textbf{63.62} & \cg \textbf{0.242} \\
    \midrule
    \multirow{7}{*}{Motion Deblurring}
    & DAPS & 28.96 & 0.754 & 31.71 & 0.116 & 23.13 & 0.572 & 55.48 & 0.187 \\ 
    & RLSD & 21.67 & 0.631 & 99.00 & 0.396 & 18.17 & 0.431 & 136.3 & 0.538 \\
    & FlowDPS & 25.98 & 0.704 & 48.88 & 0.229 & 20.91 & 0.514 & 81.88 & 0.325 \\
    & FlowChef & 20.52 & 0.626 & 69.13 & 0.388 & 19.84 & 0.528 & 128.0 & 0.431 \\
    & FLAIR & 28.50 & 0.744 & \underline{22.62} & 0.113 & 22.95 & 0.579 & \underline{43.70} & \underline{0.174} \\
    & Flower & \textbf{31.00} & \textbf{0.834} & 67.56 & \underline{0.087} & 23.51 & 0.624 & 68.60 & 0.207 \\
    & \cg Ours & \cg \underline{29.81} & \cg \underline{0.774} & \cg \textbf{21.15} & \cg \textbf{0.067} & \cg \textbf{24.29} & \cg \textbf{0.642} & \cg \textbf{36.02} & \cg \textbf{0.110} \\
    \midrule
    \multirow{7}{*}{Super-Resolution ($\times12$)}
    & DAPS & 24.54 & 0.535 & 81.93 & 0.309 & 19.13 & 0.347 & 106.9 & 0.379 \\ 
    & RLSD & 19.53 & 0.664 & 125.0 & 0.424 & 17.04 & 0.453 & 172.0 & 0.553 \\
    & FlowDPS & 26.68 & 0.725 & 36.07 & \underline{0.171} & 21.77 & 0.543 & \underline{67.01} & \underline{0.272} \\
    & FlowChef & 26.59 & 0.721 & 58.39 & 0.208 & \underline{21.86} & \underline{0.563} & 96.32 & 0.305 \\
    & FLAIR & 26.82 & \underline{0.739} & \textbf{31.33} & 0.181 & 20.42 & 0.475 & 67.91 & 0.285 \\
    & Flower & \textbf{28.10} & \textbf{0.769} & 49.55 & 0.179 & \textbf{22.16} & \textbf{0.567} & 75.64 & 0.280 \\
    & \cg Ours & \cg \underline{27.12} & \cg 0.716 & \cg \underline{34.54} & \cg \textbf{0.158} & \cg 21.16 & \cg 0.505 & \cg \textbf{65.52} & \cg \textbf{0.267} \\
    \bottomrule
  \end{tabular}
  }
\end{table*}

\vspace{-0.1cm}
\section{Experimental Results}
\vspace{-0.1cm}
\subsection{Experimental Setup}
\paragraph{Datasets and Evaluation Metrics.}
We conducted comprehensive evaluations on two distinct datasets: 1k images from the FFHQ dataset and 0.8k images from the DIV2K dataset. All experiments were performed at $768 \times 768$ resolution by resizing the original images. 
Our evaluation employs dual assessment: PSNR and SSIM as distortion metrics for pixel-level reconstruction fidelity, complemented by FID and LPIPS for perceptual quality and naturalness.

\paragraph{Baselines.}
We compare FlowLPS with two diffusion-based solvers, RLSD~\citep{zilberstein2024repulsive} and DAPS~\citep{zhang2024improving}, and four flow-based solvers, FlowChef~\citep{patel2024steering}, FlowDPS~\citep{kim2025flowdps}, FLAIR~\citep{erbach2025flair}, and Flower~\citep{pourya2025flower}. 
To ensure a matched-backbone comparison, we adapt all baselines to the same SD3 backbone used by FlowLPS. 
Unless otherwise stated, all methods are evaluated under a fixed budget of 40 NFEs.
For methods that perform repeated measurement updates within each reverse step, we allow up to 15 inner updates per timestep for a matched optimization budget.
For DAPS, each outer update requires multiple model evaluations due to its denoised-prediction ODE step, so we use 20 outer steps with 2 inner steps to match the 40-NFE budget.
Additional implementation details are provided in Appendix~\ref{sec:baseline_details}, and runtime comparisons under the matched-budget setting are reported in Appendix~\ref{sec:additional_baseline_comparisons}.

\noindent\textbf{Task Setting.}
We used the text prompt "a high quality photo of a face" for the FFHQ dataset~\citep{karras2019style} and image descriptions generated by DAPE~\cite{wu2024seesr} for DIV2K~\citep{agustsson2017ntire}. We evaluated all methods on five linear tasks: Gaussian deblurring with kernel size $181$ and intensity $9.0$, Motion deblurring with kernel size $183$ and intensity $0.5$, Super-resolution with bicubic interpolation at 12x scaling factor, Box Inpainting with a $384\times384$ rectangular mask for FFHQ and two $192\times192$ rectangular masks for DIV2K, and Random Inpainting masking 70\% of the image. All measurement operators incorporate additional Gaussian noise with $\sigma_n = 0.03$.

\begin{figure*}[t]
    \centering
    \includegraphics[width=0.97\linewidth]{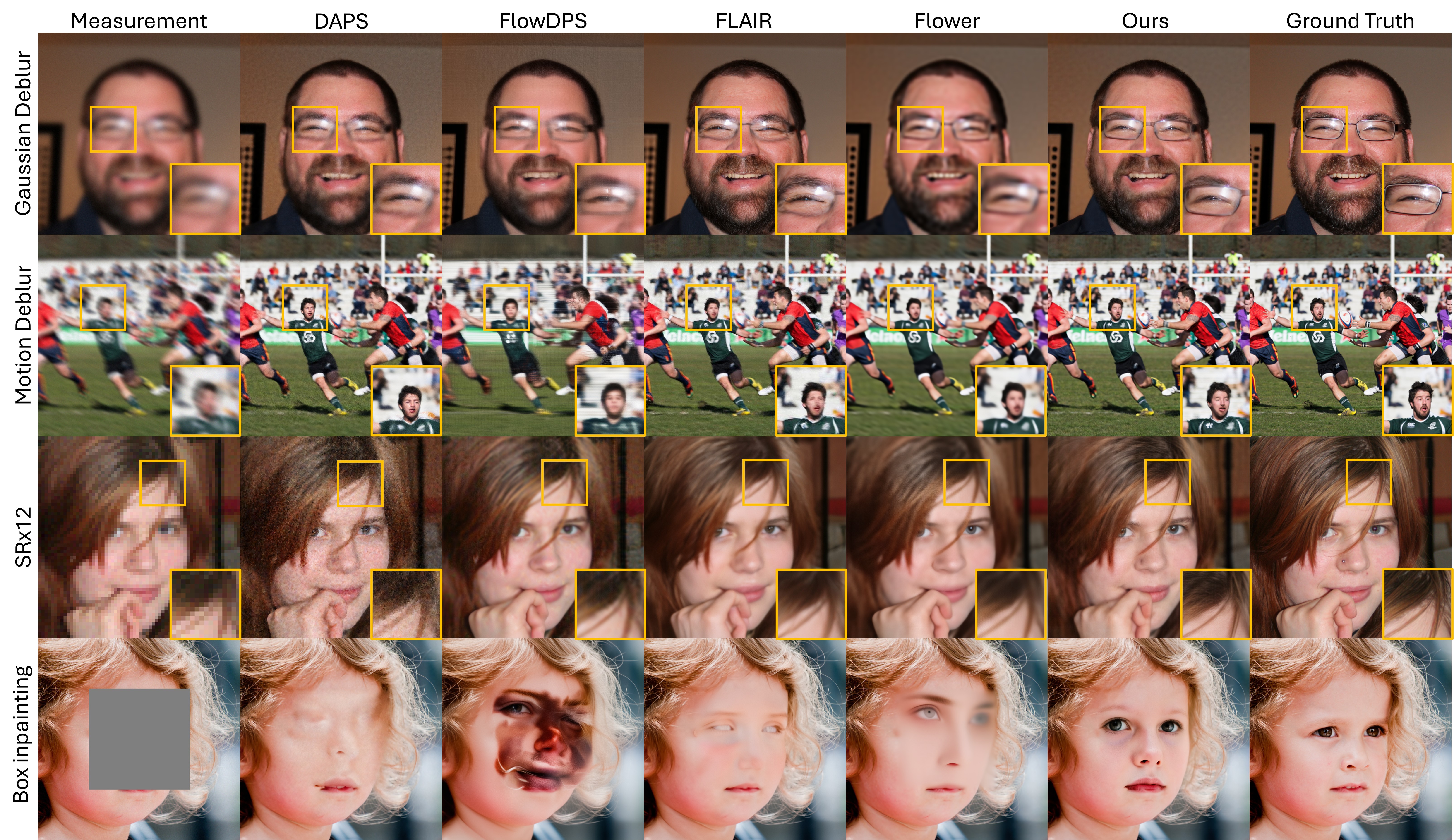}
    \caption{\textbf{Qualitative comparison on FFHQ and DIV2K datasets.} }
    \label{fig:qualitative}
    \vspace{-1.5em}
\end{figure*}

\subsection{Results}
\label{sec:main_results}

Table~\ref{whole results} summarizes the quantitative comparison across linear inverse problems.
Overall, FlowLPS achieves a favorable balance between distortion metrics and perceptual realism.
This balance is important in ill-posed inverse problems, where multiple plausible clean images can be consistent with the same measurement; therefore, distortion metrics such as PSNR/SSIM and perceptual metrics such as FID/LPIPS should be interpreted together.

Refinement-dominant solvers, such as Flower, often achieve competitive distortion metrics by aggressively reducing the measurement-consistency loss.
However, consistent with the perception--distortion trade-off~\citep{ledig2017photo, zhang2018unreasonable}, this behavior can come at the cost of perceptual fidelity, producing overly smooth or less natural textures in ambiguous regions (Fig.~\ref{fig:qualitative}).
Conversely, sampling-dominant solvers such as DAPS preserve stochastic posterior exploration, but under matched inference steps they struggle to concentrate into sharp, measurement-consistent reconstructions, limiting both distortion and perceptual metrics.

FlowLPS targets the intermediate regime.
The Langevin phase provides stochastic posterior-oriented initialization, while the proximal phase locally improves measurement consistency from that initialization.
Across tasks, FlowLPS maintains competitive PSNR/SSIM while improving perceptual metrics, indicating a better exploration--refinement balance.
The qualitative comparisons in Fig.~\ref{fig:qualitative} show the same trend: FlowLPS preserves sharper textures and more natural local details while maintaining measurement consistency.
Additional qualitative results are provided in Appendix~\ref{additional_results}.
We further evaluate FlowLPS beyond the main latent-space linear setting, including pixel-space inverse problems and phase retrieval, in Appendix~\ref{sec:pixel_space_comparisons}.

\begin{figure*}[t]
    \centering
    \includegraphics[width=0.97\linewidth]{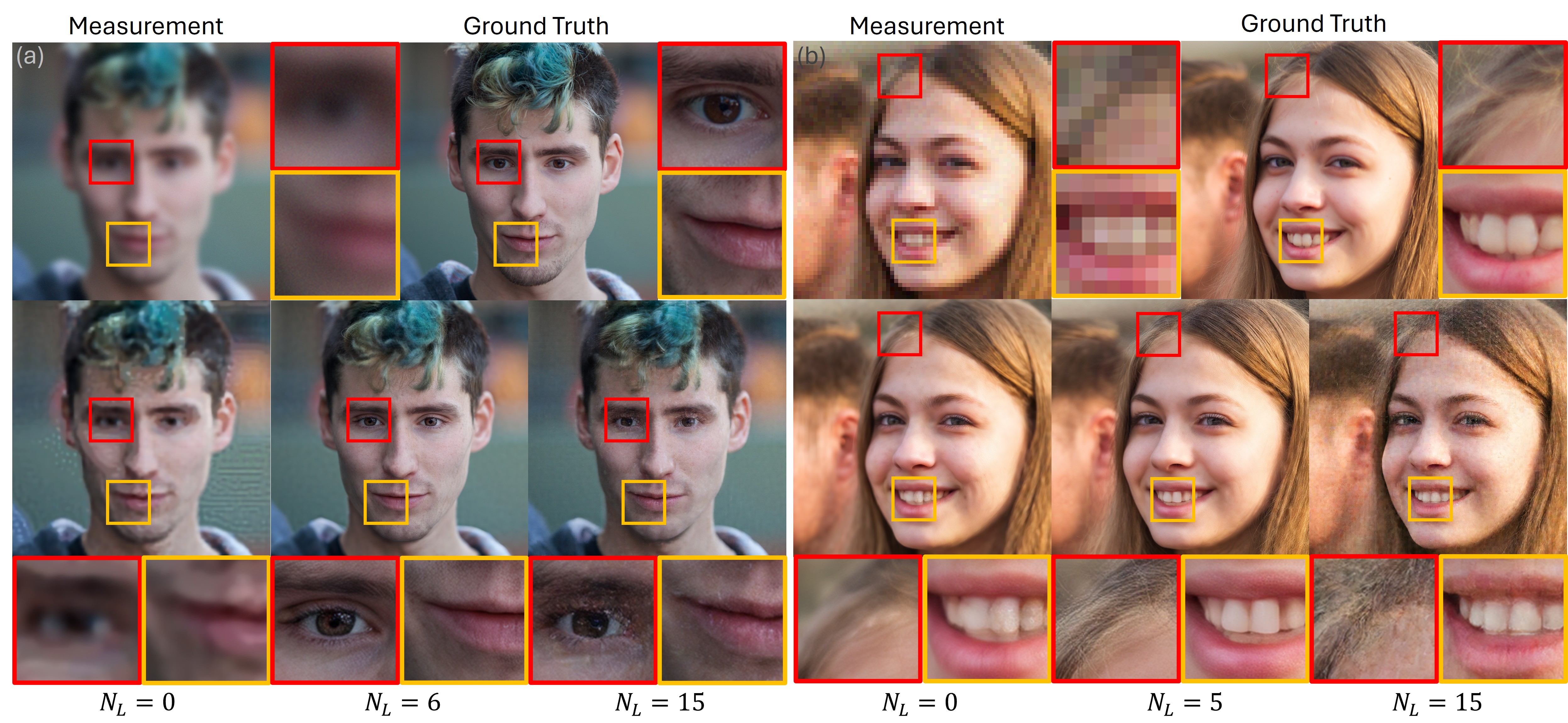}
    \caption{\textbf{Qualitative effect of Langevin steps ($N_L$).}
(a) Gaussian deblurring and (b) $\times12$ super-resolution.
With no Langevin updates ($N_L=0$), the solver behaves as a refinement-dominant method and tends to produce overly smooth reconstructions.
With excessive Langevin updates ($N_L=15$), too much budget is spent on stochastic exploration, leaving insufficient refinement and causing unstable details. Moderate values ($N_L=5$--$7$) provide the best exploration--refinement balance.}
    \label{fig:Ablation_Langevin}
    \vspace{-1em}
\end{figure*}

\subsection{Ablation Studies}
\label{langevin_effect}

\subsubsection{Number of Langevin Dynamics Steps ($N_L$)}

We vary the number of Langevin steps $N_L$ on 200 FFHQ images while keeping the total number of Langevin-Proximal update steps fixed at $N_T=15$.
Thus, increasing $N_L$ shifts computation from local proximal refinement to posterior-oriented Langevin exploration. This directly tests the update-allocation trade-off suggested by Sec.~\ref{sec:local_analysis}.

\begin{wrapfigure}{r}{0.55\textwidth}
\vspace{-0.54cm}
    \includegraphics[width=1.05\linewidth]{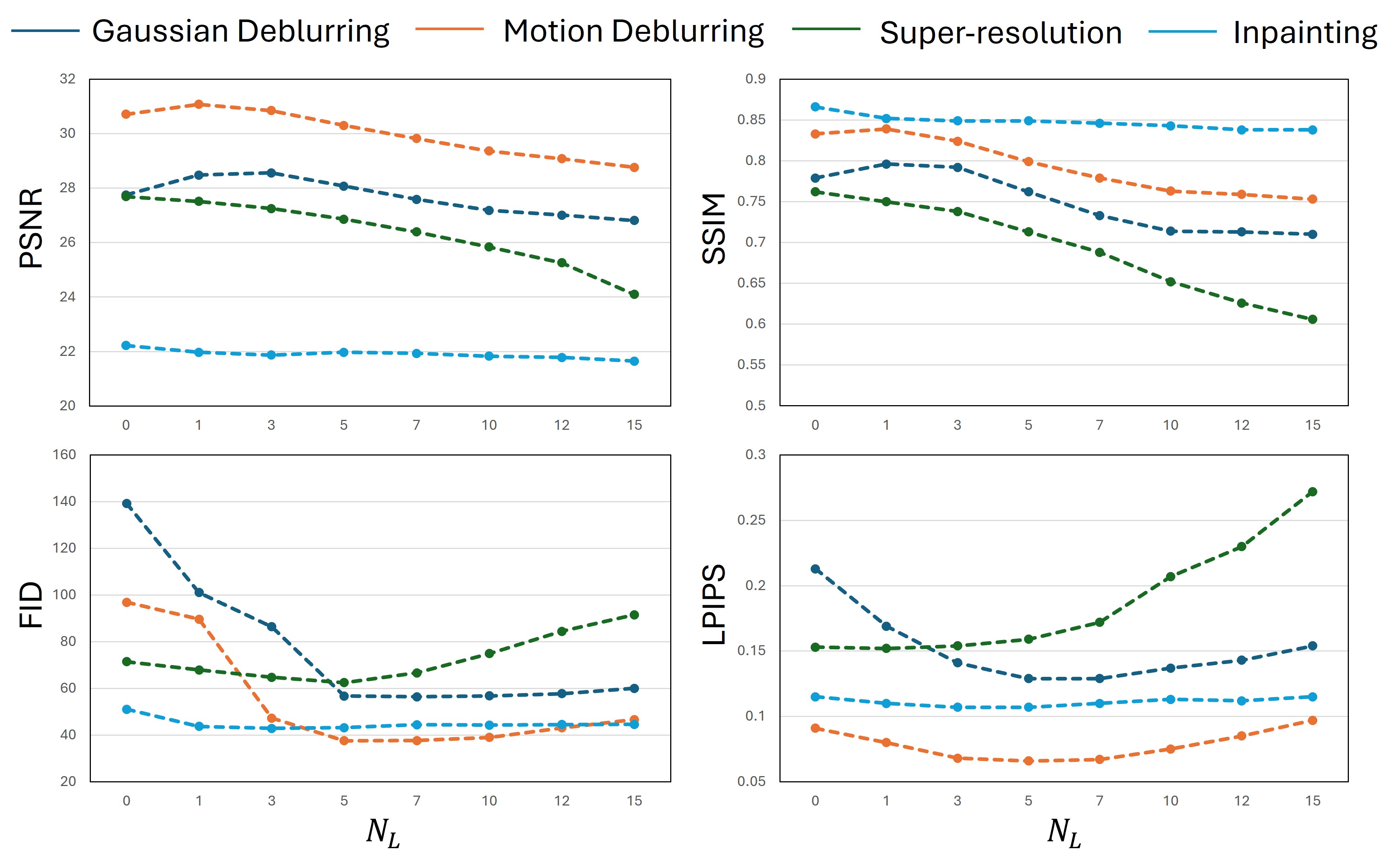}
    \caption{\textbf{Quantitative impact of Langevin dynamics steps ($N_L$).}}
    \label{fig:ablation_langevin_graph}
    \vspace{-0.5cm}
\end{wrapfigure}

Figure~\ref{fig:ablation_langevin_graph} shows a clear intermediate optimum.
When $N_L=0$, FlowLPS behaves as a refinement-dominant update initialized from the model prediction and often produces overly smooth reconstructions.
Moderate Langevin steps ($N_L=5$--$7$) improve perceptual metrics while maintaining competitive distortion metrics, suggesting that stochastic posterior-oriented initialization provides a better starting point for local refinement.
When $N_L$ is too large, the method shifts toward a sampling-dominant regime, leaving insufficient refinement to concentrate the estimate and enforce measurement consistency.

These results support the intended exploration--refinement mechanism of FlowLPS: moderate Langevin updates improve the initialization, while the proximal phase concentrates the estimate into a measurement-consistent reconstruction under the fixed update budget.
Figure~\ref{fig:Ablation_Langevin} visualizes the same behavior, and additional results are provided in Appendix~\ref{sec:N_L_ablation}.


\subsubsection{pCN Mixing Schedule and Noise Initialization}
\label{sec:rho_t strategy}

We ablate two design choices in the pCN-style re-noising step on 200 FFHQ images: the mixing schedule $\rho_t$ and the noise initialization.
Table~\ref{tab:different_rho_t} evaluates $\rho_t$ on box inpainting and shows that adaptive schedules outperform fixed ones, with $\rho_t=\sqrt{1-t}$ giving the best FID and LPIPS.
This suggests that the stochastic refresh should depend on the remaining reverse-time uncertainty, rather than being fixed across timesteps.
Using this single-step schedule, Table~\ref{table:noise_initialization} shows that initializing the refresh from the model-predicted noise improves perceptual metrics over independent Gaussian initialization.
This supports our design choice of retaining trajectory coherence while injecting controlled stochasticity.
Additional ablations on the number of pCN steps, decaying-$N_L$ schedule, and component-wise effects are provided in Appendix~\ref{num_of_pCN_steps}, Appendix~\ref{sec:decaying_NL}, and Appendix~\ref{sec:component_ablation}.
\begin{figure}[h]
\vspace{-0.1cm}
    \centering
    \begin{minipage}[h]{0.46\linewidth}\vspace{0pt}
        \captionof{table}{\textbf{pCN mixing schedule ablation.}}
          \label{tab:different_rho_t}
          \centering
          \scriptsize
          \setlength{\heavyrulewidth}{0.3pt}
        \setlength{\lightrulewidth}{0.2pt}
          \resizebox{\linewidth}{!}{
          \begin{tabular}{c|cccc}
            \toprule
            \textbf{$\rho_t$} & \textbf{PSNR} ($\uparrow$) & \textbf{SSIM} ($\uparrow$) & \textbf{FID} ($\downarrow$) & \textbf{LPIPS} ($\downarrow$) \\
            \midrule
            0 & \textbf{22.29} & 0.843 & 54.21 & 0.116\\
            0.5 & 21.47 & 0.844 & 53.79 & 0.111\\
            $1-t$ & 21.79 & 0.842 & 43.56 & 0.108\\
            $\sqrt{1-t}$ & 21.89 & \textbf{0.854} & \textbf{37.40} & \textbf{0.101} \\
            \bottomrule
          \end{tabular}
        }
    \end{minipage}\hfill
    \begin{minipage}[h]{0.51\linewidth}\vspace{0pt}
          \captionof{table}{\textbf{Noise initialization ablation.}}
          \label{table:noise_initialization}
          \centering
          \scriptsize
          \setlength{\heavyrulewidth}{0.3pt}
        \setlength{\lightrulewidth}{0.2pt}
          \resizebox{\linewidth}{!}{
          \begin{tabular}{c|c|cccc}
            \toprule
            \textbf{Task} & \textbf{Initialization}  & \textbf{PSNR} ($\uparrow$) & \textbf{FID} ($\downarrow$) & \textbf{LPIPS} ($\downarrow$) \\
            \midrule
            Super-Resolution & $\mathcal{N}(\mathbf{0},\mathbf{I})$ & \textbf{27.21} & 73.43 & 0.181\\
            ($\times12$) & Model prediction & 27.13 & \textbf{63.08} & \textbf{0.155} \\
            \midrule
            Inpainting & $\mathcal{N}(\mathbf{0},\mathbf{I})$ & \textbf{22.42} & 48.69 & 0.106 \\
            (Box) & Model prediction & 21.89 & \textbf{37.40} & \textbf{0.101} \\
            \bottomrule
          \end{tabular}
        }
    \end{minipage}
    \vspace{-0.5cm}
\end{figure}

\section{Conclusion}
\vspace{-0.2cm}
We introduced FlowLPS, a training-free latent flow inverse solver that combines posterior-oriented Langevin initialization with local measurement-consistent proximal refinement.
FlowLPS targets the practical finite-step trade-off between stochastic exploration and deterministic refinement, with a local analysis and ablations supporting this design.
Together with controlled pCN-style re-noising, FlowLPS achieves a strong balance between measurement fidelity and perceptual realism across diverse inverse problems.
\vspace{-0.1cm}
\paragraph{Limitations.}
\label{sec:limitations}
FlowLPS depends on the pretrained backbone prior, so highly out-of-distribution inputs may produce artifacts or hallucinated details.
Caution is needed in high-stakes applications such as medical, forensic, or scientific imaging.

\clearpage
\bibliographystyle{unsrtnat}
\bibliography{main}

\clearpage
\appendix
\section{Proofs}
\label{proofs}

\mapprox*
\begin{proof}
    By Bayes' rule and the Markov property, 
    \begin{align*}
        p(\vy|\vz_0,\vz_t)&=p(\vy|\vz_0),\\
        p(\vz_0|\vz_t,\vy)&=\frac{p(\vy|\vz_0,\vz_t)p(\vz_0|\vz_t)}{p(\vy|\vz_t)}\\
        &\propto p(\vy|\vz_0)p(\vz_0|\vz_t).
    \end{align*}
    Substituting the Gaussian forms gives
    \begin{align*}
        \max_{\vz_0} \log p(\vz_0|\vz_t,\vy) &= \max_{\vz_0} \log p(\vy|\vz_0)+\log p(\vz_0|\vz_t) \\
        &= \max_{\vz_0} -\frac{\|\vy-\mathcal{A}(\vz_0)\|^2}{2\sigma_n^2} -\frac{\|\vz_0-\hat{\vz}_{0|t}\|^2}{2s_t^2} \\
        &=\min_{\vz_0} \|\vy-\mathcal{A}(\vz_0)\|^2 +\frac{\sigma_n^2}{s_t^2}\|\vz_0-\hat{\vz}_{0|t}\|^2 
    \end{align*}
\end{proof}

\captureexpansion*
\begin{proof}
Let $T(\vz)=\vz-\tau\nabla\Phi(\vz)$.
Since $\Phi$ is twice differentiable and satisfies
\begin{equation}
    \mu \mathbf{I}\preceq \nabla^2\Phi(\vz)\preceq L\mathbf{I}
\end{equation}
on $\Omega_B$, the gradient map
\begin{equation}
    T(\vz)=\vz-\tau\nabla\Phi(\vz)
\end{equation}
is contractive on $\Omega_B$ for any $0<\tau\le 1/L$.
Indeed, for any $\vz\in\Omega_B$, using $\nabla\Phi(\vz_B^\star)=0$, we have
\begin{align}
    T(\vz)-\vz_B^\star
    &=
    \vz-\vz_B^\star
    -
    \tau\left(\nabla\Phi(\vz)-\nabla\Phi(\vz_B^\star)\right) \\
    &=
    \left[
    \mathbf{I}
    -
    \tau
    \int_0^1
    \nabla^2\Phi\bigl(\vz_B^\star+s(\vz-\vz_B^\star)\bigr)
    \,ds
    \right]
    (\vz-\vz_B^\star).
\end{align}
The eigenvalues of the averaged Hessian lie in $[\mu,L]$.
Therefore, since $0<\tau\le 1/L$, the operator norm of the bracketed matrix is at most
\begin{equation}
    q=1-\tau\mu.
\end{equation}
Thus,
\begin{equation}
    \|T(\vz)-\vz_B^\star\|
    \le
    q\|\vz-\vz_B^\star\|.
\end{equation}
Applying this contraction recursively for $N$ refinement steps gives
\begin{equation}
    \|R_N(\vz)-\vz_B^\star\|
    \le
    q^N\|\vz-\vz_B^\star\|.
\end{equation}
This proves the claim.
\end{proof}

\ulanoisefloor*
\begin{proof}
Let
\begin{equation}
    T(\vz)=\vz-\tau\nabla\Phi(\vz).
\end{equation}
From Proposition~\ref{prop:capture_expansion}, $T$ is locally contractive:
\begin{equation}
    \|T(\vz)-\vz_B^\star\|
    \le
    q\|\vz-\vz_B^\star\|.
\end{equation}
Conditioned on $\vz_k$, the ULA update satisfies
\begin{equation}
    \vz_{k+1}-\vz_B^\star
    =
    T(\vz_k)-\vz_B^\star
    +
    \sqrt{2\tau}\boldsymbol{\xi}_k.
\end{equation}
Taking conditional expectation and using
$\mathbb{E}[\boldsymbol{\xi}_k]=0$ and
$\mathbb{E}\|\boldsymbol{\xi}_k\|^2=d$, we obtain
\begin{align}
    \mathbb{E}
    \left[
        \|\vz_{k+1}-\vz_B^\star\|^2
        \mid \vz_k
    \right]
    &=
    \|T(\vz_k)-\vz_B^\star\|^2
    +
    2\tau d \\
    &\le
    q^2
    \|\vz_k-\vz_B^\star\|^2
    +
    2\tau d.
\end{align}
Taking total expectation gives the recursion
\begin{equation}
    a_{k+1}\le q^2 a_k+2\tau d,
    \qquad
    a_k=\mathbb{E}\|\vz_k-\vz_B^\star\|^2.
\end{equation}
Unrolling the recursion yields
\begin{equation}
    a_N
    \le
    q^{2N}a_0
    +
    2\tau d
    \sum_{\ell=0}^{N-1}q^{2\ell}
    =
    q^{2N}a_0
    +
    2\tau d
    \frac{1-q^{2N}}{1-q^2}.
\end{equation}
This proves the result.
\end{proof}

\section{Implementation Details}
\label{implementation_details}
In this section, we provide implementation details for FlowLPS and all baseline methods.
Unless otherwise stated, the main experiments use Stable Diffusion 3.5-Medium~\citep{esser2024scaling} as the matched backbone for all baselines and FlowLPS.
All experiments were run on a single NVIDIA RTX 3090 GPU.
We additionally report original SD2.1 diffusion-backbone~\citep{patil2022stable} results for DAPS and RLSD in Sec.~\ref{sec:additional_baseline_comparisons}.

\subsection{FlowLPS Configuration}
\label{sec:configuration}
We employed a fixed number of Langevin-Proximal update steps, $N_T = 15$ for our main experiments. The variance parameter $s_t^2=t$ was used in the proximal objective. A constant step size $\zeta_t = 10^{-4}$ was used for Langevin dynamics across all timesteps. For the proximal optimization, we utilized the SGD optimizer. We used an initial learning rate of 0.5 for super-resolution and 0.1 for other tasks. To ensure stability during optimization, we applied a step-based learning rate decay schedule: for super-resolution, the learning rate is decayed by a factor of 0.85 every 5 iterations, and for inpainting, by 0.65 every 10 iterations. For the forward measurement operators, we adopted the implementations from DPS~\citep{chung2023diffusion}. Based on empirical findings that negative prompts improve results, we used the negative prompt "blurry, noisy, a bad quality photo" for classifier-free guidance. All experiments were conducted with a Classifier-Free Guidance (CFG) scale of $2.0$.

\begin{figure}[!h]
    \centering
    \includegraphics[width=0.75\linewidth]{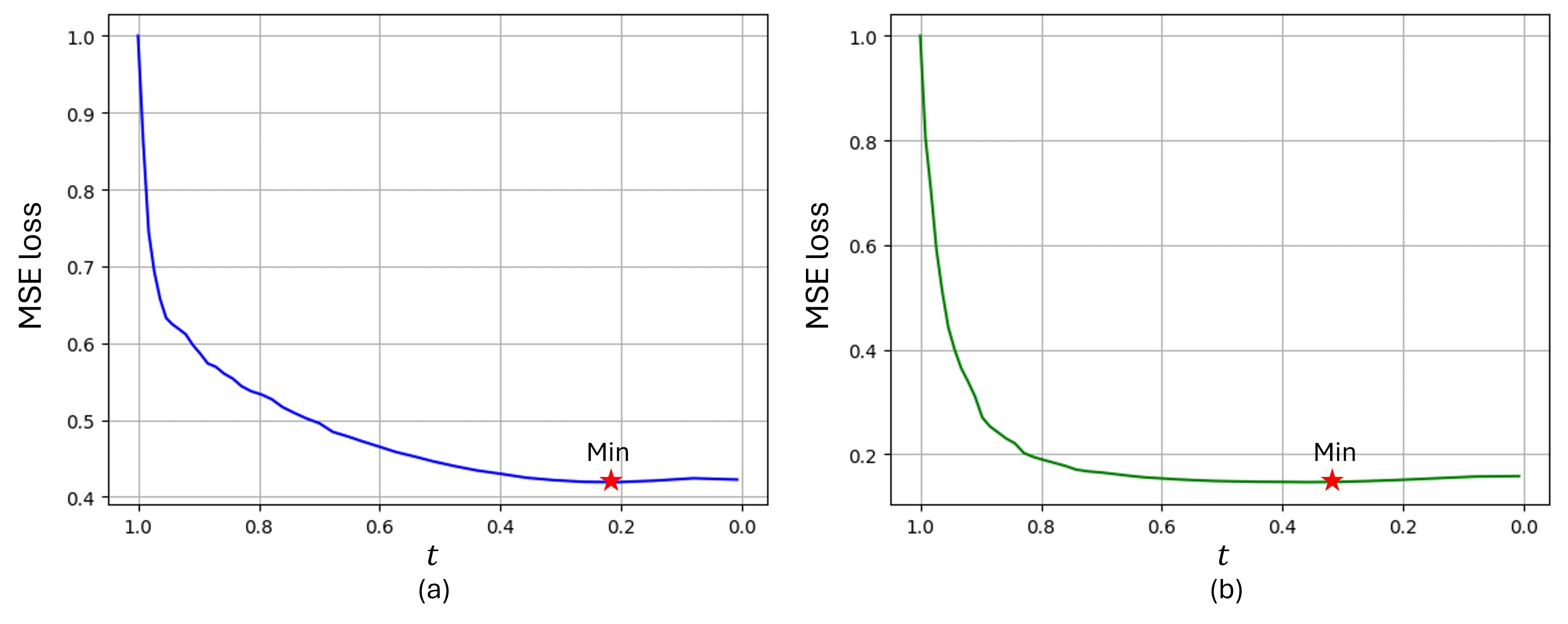}
    \caption{\textbf{Reconstruction MSE error vs. Timestep $t$.} (a) Deblurring, (b) Super-resolution. The plot shows the relative MSE of the estimated clean image $\mathcal{D}(\vz_{0|t}^*)$ for deblurring (averaged over Gaussian/Motion) and super-resolution. The error minimizes around $t \approx 0.2$ for deblurring and $t \approx 0.3$ for super-resolution, justifying the use of task-specific truncated time schedules.}
    \label{fig:timesteps}
\end{figure}

\noindent\textbf{Time Schedule.} 
Fig.~\ref{fig:timesteps} illustrates the Mean Squared Error (MSE) loss of the reconstructed clean image estimate $\mathcal{D}(\vz_{0|t}^*)$ as a function of timestep $t$, scaled relative to the maximum value. The results are averaged over 100 FFHQ images, for both super-resolution and deblurring tasks, with the deblurring curve representing the average of both Gaussian and motion deblurring tasks. We observed that the MSE decreases as $t$ decreases, reaching a minimum around $t \approx 0.2$ for deblurring and $t \approx 0.3$ for super-resolution, before beginning to rise as $t \to 0$. Based on this trend, we adopted a truncated time schedule to stop the reverse process before significant divergence occurs. Specifically, we set the sampling schedule start point to $40+\alpha$ (using the first 40 steps). We set $\alpha=5$ for super-resolution and $\alpha=3$ for other tasks, which align with the optimal stopping points observed in our analysis. We found these settings empirically provide the best trade-off between reconstruction fidelity (MSE) and perceptual details.

A detailed summary of hyperparameters for all tasks is provided in Table~\ref{hyperparameters}.

\begin{table}[!h]
  \caption{\textbf{Hyperparameters for FlowLPS experiments.} $N_L$ denotes the number of Langevin steps, $N_P$ denotes the number of Proximal optimization steps, $N_T$ denotes the total number of Langevin-Proximal optimization steps, $\zeta_t$ denotes the step size of Langevin dynamics, and $\alpha$ denotes the sampling schedule offset parameter.}
  \vspace{0.5em}
  \label{hyperparameters}
  \centering
  \scriptsize
  \resizebox{0.7\linewidth}{!}{
  \begin{tabular}{c|ccccc}
    \toprule
     & \shortstack{Inpainting \\ (Box)} & \shortstack{Inpainting \\ (Random)} & \shortstack{Gaussian \\ Deblurring} & \shortstack{Motion \\ Deblurring} & \shortstack{Super-resolution \\ ($\times12$)}  \\
    \midrule
    $\zeta_t$ & $10^{-4}$ & $10^{-4}$ & $10^{-4}$ & $10^{-4}$ & $10^{-4}$\\
    $\alpha$ & $3$ & $3$ & $3$ & $3$ & $5$\\
    $N_L$ & $4$ & $5$ & $6$ & $6$ & $4$\\
    $N_P$ & $11$ & $10$ & $9$ & $9$ & $11$\\
    $N_T$ & $15$ & $15$ & $15$ & $15$ & $15$\\
    \bottomrule
  \end{tabular}
}
\end{table}

\subsection{Baseline Details}
\label{sec:baseline_details}

\noindent\textbf{DAPS~\citep{zhang2024improving}.}
We adapt DAPS to the SD3 backbone and use 20 annealing steps with 2 inner denoised-prediction ODE evaluations, resulting in 40 NFEs in total.
We use a CFG scale of 2.0 for all tasks.
For most tasks, we use 15 MCMC updates per annealing step with a learning rate of $10^{-4}$.
For super-resolution, we found that 15 MCMC updates produced degenerate reconstructions under the SD3 adaptation; therefore, we use 60 MCMC updates with a learning rate of $10^{-3}$ and decay the learning rate by a factor of 0.7 every 5 MCMC iterations.
All hyperparameters were selected by grid search on a validation set of 200 images.

\noindent\textbf{RLSD~\citep{zilberstein2024repulsive}.}
We adapt RLSD to the SD3 backbone and use 40 NFEs.
We do not use pixel-space augmentation, as we found that it degraded performance, consistent with observations in prior latent-space inverse solvers~\citep{erbach2025flair}.
We also disable the repulsive term, since it reduced reconstruction fidelity in our experiments, consistent with the original findings.
We set $(lr_z,w_t)=(0.1,1.5)$ for deblurring tasks, $(0.15,1.5)$ for super-resolution, and $(0.2,1.2t)$ for inpainting tasks.

\noindent\textbf{FlowDPS~\citep{kim2025flowdps}.}
We use 40 NFEs with 15 optimization steps per iteration and a CFG scale of 2.0.
The step size is set to 8.0 for super-resolution, Gaussian deblurring, and motion deblurring, and 12.0 for inpainting tasks.

\noindent\textbf{FlowChef~\citep{patel2024steering}.}
We use 40 NFEs with a single optimization step per iteration.
Although we attempted to increase the number of optimization steps to 15 for a matched inner-update budget, this did not improve performance.
We therefore use the original single-step update with a step size of 1.5 across all tasks.

\noindent\textbf{FLAIR~\citep{erbach2025flair}.}
We configure FLAIR with 40 NFEs and 15 optimization steps per iteration.
We use a CFG scale of 2.0 and follow the default settings, except for super-resolution where the data-term optimizer learning rate is set to 1.0.

\noindent\textbf{Flower~\citep{pourya2025flower}.}
We use 40 NFEs, 15 optimization steps per iteration, and a CFG scale of 2.0.
Consistent with the authors' observations, we found that the optional uncertainty term destabilized restoration; thus, we set $\gamma=0$.

\subsection{Additional Baseline Comparisons}
\label{sec:additional_baseline_comparisons}

\paragraph{Original diffusion-backbone results.}
Although the main experiments adapt all baselines to the matched SD3 backbone, we additionally report DAPS and RLSD using their original SD2.1 diffusion backbone for reference.
For the original DAPS setting, we use 5 ODE steps and 50 annealing steps, resulting in 250 NFEs.
Hyperparameters were selected by grid search on a validation set of 200 images; for example, we use 50 MCMC steps with learning rate $10^{-6}$ and CFG scale 6.0 for Gaussian deblurring, and 50 MCMC steps with learning rate $10^{-5}$ and CFG scale 5.0 for super-resolution.
For the original RLSD setting, we use 250 NFEs, disable the repulsive term, and set the CFG scale to 6.0 for all tasks.
Table~\ref{tab:original_backbone_results} reports DAPS and RLSD under their original SD2.1 diffusion-backbone protocols, together with our SD3 adaptations and FlowLPS.
These results are provided for reference, since backbone choice and inference budget can substantially affect performance.
Our main claims are based on the matched SD3 setting, while this table provides additional context on backbone-dependent performance variations.

\begin{table*}[t]
  \caption{\textbf{Additional comparison of original diffusion backbones and SD3 adaptations.} 
We report DAPS and RLSD using their original SD2.1 diffusion backbone and our SD3 adaptation, together with FlowLPS.}
  \tiny
  \label{tab:original_backbone_results}
  \centering
  \resizebox{\linewidth}{!}{
  \begin{tabular}{l|l|cccc|cccc}
    \toprule
    \multirow{2}{*}{\textbf{Task}} & \multirow{2}{*}{\textbf{Method}} & \multicolumn{4}{c|}{\textbf{FFHQ 1k}} & \multicolumn{4}{c}{\textbf{DIV2K 0.8k}} \\
    & & \textbf{PSNR} ($\uparrow$) & \textbf{SSIM} ($\uparrow$) & \textbf{FID} ($\downarrow$) & \textbf{LPIPS} ($\downarrow$) & \textbf{PSNR} ($\uparrow$) & \textbf{SSIM} ($\uparrow$) & \textbf{FID} ($\downarrow$) & \textbf{LPIPS} ($\downarrow$) \\
    \midrule
    \multirow{5}{*}{Inpainting (Box)} 
    & DAPS (SD2) & 19.77 & 0.846 & 53.22 & 0.139 & 22.50 & 0.807 & 30.57 & 0.091 \\ 
    & DAPS (SD3) & \underline{21.78} & 0.842 & 24.75 & \underline{0.114} & 23.72 & 0.841 & 19.96 & 0.080 \\ 
    & RLSD (SD2) & 13.08 & 0.766 & 151.5 & 0.261 & 17.11 & 0.822 & 75.41 & 0.148 \\
    & RLSD (SD3) & 16.32 & 0.519 & 191.6 & 0.538 & 16.25 & 0.417 & 164.1 & 0.535 \\
    & Ours & 21.68 & \underline{0.849} & \textbf{21.90} & \textbf{0.107} & 23.73 & 0.862 & \textbf{15.50} & \textbf{0.065} \\
    \midrule
    \multirow{5}{*}{Inpainting (Random)} 
    & DAPS (SD2) & 30.99 & \underline{0.884} & 16.48 & \underline{0.035} & 25.53 & \underline{0.764} & \underline{24.17} & 0.064 \\ 
    & DAPS (SD3) & \underline{33.41} & \underline{0.857} & \underline{13.28} & \underline{0.034} & 27.97 & 0.799 & 15.37 & 0.039 \\ 
    & RLSD (SD2) & 17.24 & 0.383 & 116.6 & 0.405 & 16.98 & 0.396 & 107.0 & 0.382 \\
    & RLSD (SD3) & 17.04 & 0.539 & 123.6 & 0.441 & 15.63 & 0.392 & 137.7 & 0.480 \\
    & Ours & \textbf{34.34} & \textbf{0.886} & \textbf{7.460} & \textbf{0.019} & \textbf{28.85} & \underline{0.841} & \textbf{9.360} & \textbf{0.025} \\
    \midrule
    \multirow{5}{*}{Gaussian Deblurring} 
    & DAPS (SD2) & \underline{28.20} & \underline{0.772} & 49.83 & 0.174 & \underline{21.94} & \textbf{0.550} & 75.34 & 0.286 \\ 
    & DAPS (SD3) & 27.21 & 0.715 & 41.72 & 0.171 & 21.42 & 0.498 & 81.81 & 0.294 \\ 
    & RLSD (SD2) & 25.50 & 0.674 & 88.23 & 0.227 & 21.86 & 0.538 & 104.1 & \textbf{0.242} \\
    & RLSD (SD3) & 21.42 & 0.626 & 110.3 & 0.426 & 17.90 & 0.424 & 154.9 & 0.571 \\
    & Ours & \underline{27.76} & 0.737 & \underline{31.22} & \textbf{0.130} & \textbf{21.97} & \underline{0.549} & \textbf{63.62} & \textbf{0.242} \\
    \midrule
    \multirow{5}{*}{Motion Deblurring} 
    & DAPS (SD2) & \underline{30.85} & \underline{0.830} & 43.17 & 0.095 & \textbf{24.56} & \textbf{0.661} & 49.03 & 0.136 \\ 
    & DAPS (SD3) & 28.96 & 0.754 & 31.71 & 0.116 & 23.13 & 0.572 & 55.48 & 0.187 \\ 
    & RLSD (SD2) & 25.73 & 0.719 & 92.52 & 0.269 & 22.16 & 0.578 & 87.06 & 0.260 \\
    & RLSD (SD3) & 21.67 & 0.631 & 99.00 & 0.396 & 18.17 & 0.431 & 136.3 & 0.538 \\
    & Ours & \underline{29.81} & \underline{0.774} & \textbf{21.15} & \textbf{0.067} & \underline{24.29} & \underline{0.642} & \textbf{36.02} & \textbf{0.110} \\
    \midrule
    \multirow{5}{*}{Super-Resolution ($\times12$)} 
    & DAPS (SD2) & \underline{27.61} & 0.727 & 41.55 & 0.177 & 18.02 & 0.415 & 80.84 & 0.368 \\
    & DAPS (SD3) & 24.54 & 0.535 & 81.93 & 0.309 & 19.13 & 0.347 & 106.9 & 0.379 \\
    & RLSD (SD2) & 22.18 & 0.644 & 133.5 & 0.404 & 18.81 & 0.450 & 187.3 & 0.430 \\ 
    & RLSD (SD3) & 19.53 & 0.664 & 125.0 & 0.424 & 17.04 & 0.453 & 172.0 & 0.553 \\
    & Ours & \underline{27.12} & 0.716 & \underline{34.54} & \textbf{0.158} & 21.16 & 0.505 & \textbf{65.52} & \textbf{0.267} \\
    \bottomrule
  \end{tabular}
  }
\end{table*}

\paragraph{Computational Costs.}
We report runtime comparisons for the inpainting task on an RTX 3090 GPU.
For SD3-adapted baselines, we use the matched 40-NFE protocol from the main experiments.
For the original SD2 diffusion baselines, we report their original high-budget 250-NFE protocols for reference.

RLSD (SD3) and FlowChef are faster because they use only a single measurement update per reverse step.
In contrast, FlowDPS, FLAIR, Flower, and FlowLPS perform repeated measurement updates, leading to comparable runtimes.
FlowLPS has a runtime similar to FLAIR and Flower, while achieving stronger reconstruction quality.
DAPS (SD3) is faster under the 40-NFE setting because its decoupled design performs measurement updates over fewer outer reverse steps.

\begin{table*}[h]
  \caption{\textbf{Runtime comparison on the inpainting task.}
We report seconds per image on an RTX 3090 GPU. SD3-adapted baselines follow the matched 40-NFE protocol, while original SD2 baselines use 250 NFEs.}
  \tiny
  \label{tab:computational_cost}
  \centering
  \resizebox{\linewidth}{!}{
  \begin{tabular}{l|ccccccccc}
    \toprule
    Method & DAPS (SD2) & DAPS (SD3) & RLSD (SD2) & RLSD (SD3) & FlowDPS & FlowChef & FLAIR & Flower & FlowLPS (Ours) \\
    \midrule
    Runtime (s) & 1005 & 107.6 & 332.9 & 27.46 & 207.1 & 23.06 & 288.7 & 270.8 & 264.5 \\
    NFE & 250 & 40 & 250 & 40 & 40 & 40 & 40 & 40 & 40 \\
    \bottomrule
  \end{tabular}
  }
\end{table*}

\paragraph{Higher-budget DAPS.}
Under the matched 40-NFE protocol, DAPS uses 20 outer reverse steps because each outer step requires 2 denoised-prediction ODE evaluations.
Since measurement updates are performed at each outer step, this means that DAPS applies measurement correction at only half as many reverse timesteps as other 40-step baselines.
To isolate the effect of this reduced number of measurement-correction timesteps, we additionally evaluate a higher-budget DAPS variant with 40 outer reverse steps and 3 inner denoised-prediction ODE evaluations, resulting in 120 NFEs.
The results are reported in Table~\ref{tab:higher_budget_daps}.

As shown in Table~\ref{tab:higher_budget_daps}, increasing the budget improves DAPS in several perceptual metrics, especially FID and LPIPS.
However, even with 120 NFEs, DAPS remains consistently behind FlowLPS in perceptual quality and often in pixel-level fidelity as well.
This supports our interpretation that, under practical inference budgets, Langevin-only posterior exploration struggles to consistently concentrate into sharp, measurement-consistent reconstructions.

\begin{table*}[t]
  \caption{\textbf{Higher-budget DAPS experiments on FFHQ 1k dataset.} \textbf{Bold}: best. \underline{Underline}: second best.}
  \tiny
  \label{tab:higher_budget_daps}
  \centering
  \resizebox{0.7\linewidth}{!}{
  \begin{tabular}{l|l|c|cccc}
    \toprule
    \multirow{2}{*}{\textbf{Task}} & \multirow{2}{*}{\textbf{Method}} & \multirow{2}{*}{NFE} & \multicolumn{4}{c}{\textbf{FFHQ 1k}}\\
    & & &\textbf{PSNR} ($\uparrow$) & \textbf{SSIM} ($\uparrow$) & \textbf{FID} ($\downarrow$) & \textbf{LPIPS} ($\downarrow$) \\
    \midrule
    \multirow{3}{*}{Inpainting (Box)} 
    & DAPS & 40 & \underline{21.78} & 0.842 & 24.75 & \underline{0.114} \\ 
    & DAPS & 120  & \textbf{22.10} & \underline{0.848} & \underline{24.57} & \textbf{0.107} \\ 
    & Ours & 40 & 21.68 & \textbf{0.849} & \textbf{21.90} & \textbf{0.107} \\
    \midrule
    \multirow{3}{*}{Inpainting (Random)} 
    & DAPS & 40 & \underline{33.41} & 0.857 & 13.28 & 0.034 \\ 
    & DAPS & 120 & \textbf{34.99} & \textbf{0.906} & \underline{9.811} & \underline{0.025} \\
    & Ours & 40 & \underline{34.34} & \underline{0.886} & \textbf{7.460} & \textbf{0.019} \\
    \midrule
    \multirow{3}{*}{Gaussian Deblurring} 
    & DAPS & 40 & \underline{27.21} & \underline{0.715} & 41.72 & 0.171 \\ 
    & DAPS & 120 & 27.09 & 0.710 & \underline{39.53} & \underline{0.167} \\
    & Ours & 40 & \textbf{27.76} & \textbf{0.737} & \textbf{31.22} & \textbf{0.130} \\
    \midrule
    \multirow{3}{*}{Motion Deblurring}
    & DAPS & 40 & \underline{28.96} & \underline{0.754} & 31.71 & 0.116 \\ 
    & DAPS & 120 & 28.80 & 0.747 & \underline{29.85} & \underline{0.112} \\
    & Ours & 40 & \textbf{29.81} & \textbf{0.774} & \textbf{21.15} & \textbf{0.067} \\
    \midrule
    \multirow{3}{*}{Super-Resolution ($\times12$)} 
    & DAPS & 40 & \underline{24.54} & 0.535 & 81.93 & 0.309  \\
    & DAPS & 120 & 24.25 & \underline{0.546} & \underline{71.26} & \underline{0.283} \\
    & Ours & 40 & \textbf{27.12} & \textbf{0.716} & \textbf{34.54} & \textbf{0.158} \\
    \bottomrule
  \end{tabular}
  }
\end{table*}

\begin{table*}[t]
  \caption{\textbf{Pixel-space results on CelebA-HQ 1k} with $\sigma_n=0.03$. \textbf{Bold}: best. \underline{Underline}: second best.}
  \tiny
  \label{tab:pixel_space_results}
  \centering
  \resizebox{0.7\linewidth}{!}{
  \begin{tabular}{l|l|cccc}
    \toprule
    \multirow{2}{*}{\textbf{Task}} & \multirow{2}{*}{\textbf{Method}} & \multicolumn{4}{c}{\textbf{CelebA-HQ 1k}}\\
    & & \textbf{PSNR} ($\uparrow$) & \textbf{SSIM} ($\uparrow$) & \textbf{FID} ($\downarrow$) & \textbf{LPIPS} ($\downarrow$) \\
    \midrule
    \multirow{6}{*}{Inpainting (Box)} 
    & DAPS & 24.14 & 0.775 & 41.43 & 0.128 \\ 
    & RLSD & 22.61 & 0.557 & 53.15 & 0.205 \\ 
    & FlowChef & 18.16 & 0.633 & 35.58 & 0.296 \\
    & FlowDPS & 23.08 & 0.814 & 48.81 & 0.171 \\
    & Flower & \textbf{25.66} & \textbf{0.835} & \underline{35.51} & \underline{0.107} \\
    & Ours & \underline{24.46} & \underline{0.821} & \textbf{26.91} & \textbf{0.106} \\
    \midrule
    \multirow{6}{*}{Super-Resolution (x8)} 
    & DAPS & 21.57 & 0.362 & 92.61 & 0.394 \\ 
    & RLSD & 24.05 & 0.430 & 75.57 & 0.336 \\ 
    & FlowChef & 20.89 & 0.563 & \underline{37.13} & 0.326 \\
    & FlowDPS & \underline{26.87} & \underline{0.766} & 49.19 & \textbf{0.203} \\
    & Flower & \textbf{27.35} & \textbf{0.770} & 45.57 & 0.232 \\
    & Ours & 26.05 & 0.725 & \textbf{28.61} & \underline{0.214} \\
    \midrule
    \multirow{6}{*}{Phase Retrieval} 
    & DAPS & 21.82 & 0.611 & 117.2 & 0.352 \\ 
    & RLSD & 15.35 & 0.380 & 154.5 & 0.503 \\ 
    & FlowChef & 16.61 & 0.397 & 224.8 & 0.535 \\
    & FlowDPS & 21.49 & 0.634 & 106.7 & 0.359 \\
    & Flower & \textbf{31.08} & \textbf{0.784} & \underline{69.37} & \underline{0.220} \\
    & Ours & \underline{30.75} & \underline{0.775} & \textbf{68.10} & \textbf{0.218} \\
    \bottomrule
  \end{tabular}
  }
\end{table*}

\subsection{Pixel-Space Inverse Problems and Phase Retrieval}
\label{sec:pixel_space_comparisons}

Although FlowLPS is primarily designed for latent-space flow models, its Langevin-Proximal principle is not restricted to latent representations.
We therefore evaluate FlowLPS with a pixel-space flow model~\citep{liu2023flow} on CelebA-HQ 1k.
In addition to linear inverse problems, we also evaluate phase retrieval as a representative nonlinear inverse problem.
For phase retrieval, which is sensitive to initialization due to non-uniqueness, we generate 4 reconstructions for each method and report the sample with the best PSNR.

Table~\ref{tab:pixel_space_results} shows the results.
Consistent with the latent-space experiments, Flower achieves strong pixel-level metrics but weaker perceptual scores, often leading to blurry reconstructions.
In contrast, FlowLPS achieves stronger perceptual quality while maintaining competitive pixel-level fidelity.
On phase retrieval, FlowLPS achieves the best perceptual scores and competitive distortion metrics, suggesting that the proposed Langevin-Proximal principle is not limited to linear inverse problems.

\section{Additional Ablation Studies}
\label{ablation}
\subsection{Effectiveness of the Langevin-Proximal Update}
\label{sec:N_L_ablation}
\begin{figure*}[t]
    \centering
    \includegraphics[width=\linewidth]{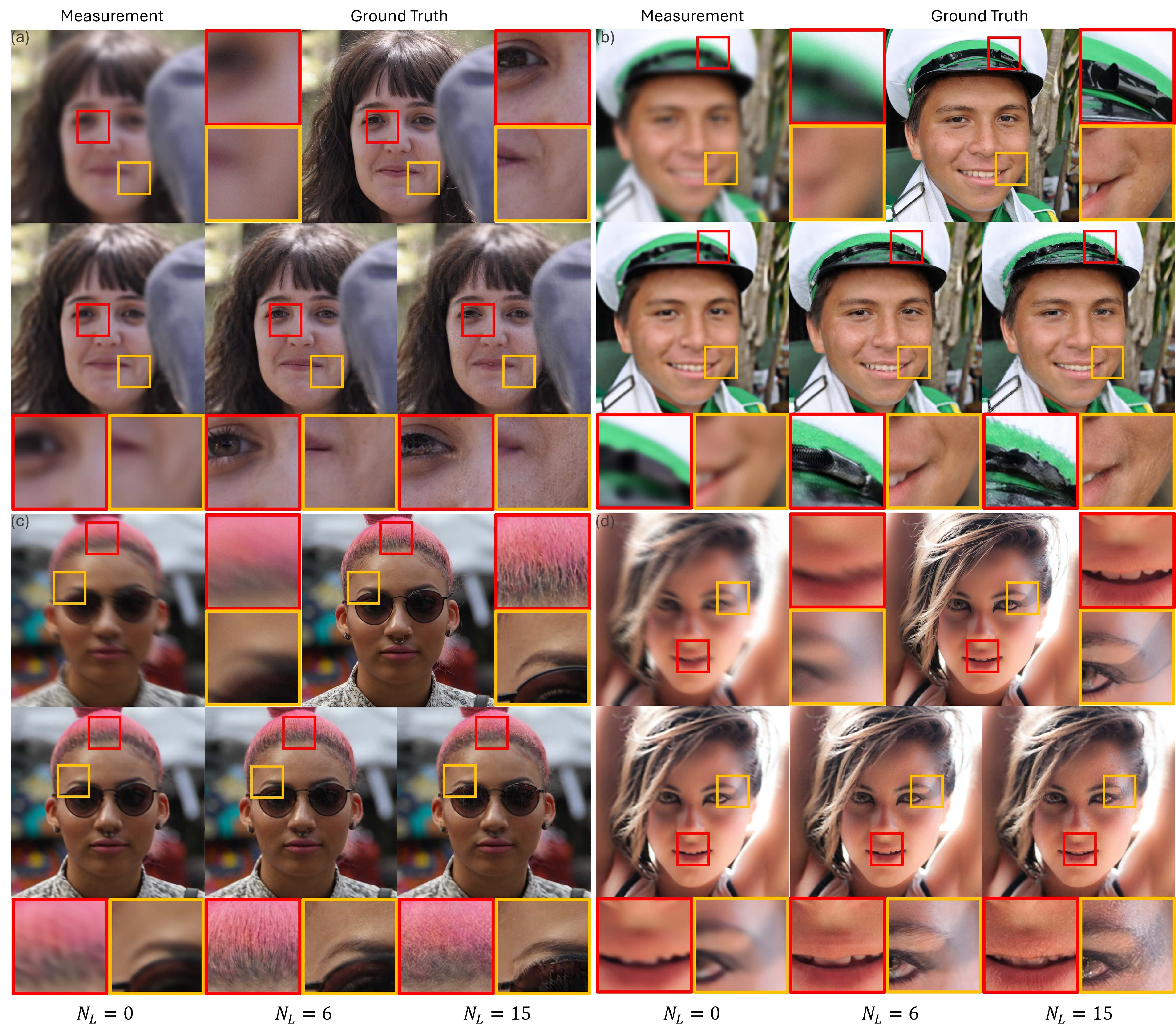}
    \caption{\textbf{Visual artifacts from excessive Langevin steps.} (a), (b) Gaussian deblurring, (c), (d) Motion deblurring. $N_L=0$ yields over-smoothed results. $N_L=15$ introduces speckled, high-frequency noise. $N_L=6$ produces clean, high-fidelity images.}
    \label{fig:grainy_image}
\end{figure*}

While we demonstrated the quantitative trade-off of Langevin steps in Section~\ref{langevin_effect} of the main paper, here we provide additional visual evidence to further clarify the impact of $N_L$.
Fig.~\ref{fig:grainy_image} compares the visual quality across different $N_L$ settings:
$N_L=0$ (pure optimization) leads to high FID/LPIPS due to blurriness. Moderate stochasticity ($N_L=5-7$) achieves an optimal balance, producing detailed restorations while maintaining measurement consistency. Conversely, excessive Langevin steps ($N_L>10$) degrade perceptual quality, introducing high-frequency artifacts and grainy noise patterns. Fig.~\ref{fig:grainy_image} visualizes these extremes, contrasting the over-smoothed textures of pure optimization ($N_L=0$) with the noisy artifacts of excessive Langevin dynamics ($N_L=15$).

\subsection{Dynamic Update-Budget Allocation}
\label{sec:decaying_NL}
In this section, we analyze whether the update budget of FlowLPS can be reduced without substantially degrading reconstruction quality.

\paragraph{Decaying Langevin Steps ($N_L$).}
We hypothesize that the benefit of Langevin updates decreases over reverse time.
In early sampling stages (high $t$), the clean estimate is less reliable, so posterior-oriented Langevin exploration is more useful.
In later stages (low $t$), the flow prediction becomes more stable, making intensive Langevin updates less necessary.

To test this, we implement a linear decay schedule for $N_L$, reducing it toward zero at the end of sampling.
Our goal is to reduce the update budget to $N_T=N_L+N_P=10$ in the final block.
For deblurring, we reduce $N_L$ from $6 \to 1$.
For super-resolution and inpainting ($4 \to 0$), we additionally remove one proximal step to match the $N_T=10$ target.
As shown in Table~\ref{langevin_decaying}, this dynamic allocation maintains performance close to the fixed schedule, suggesting that Langevin updates are most useful in early stages and can be reduced later to improve efficiency.

\begin{table}[h]
  \caption{\textbf{Performance comparison: Fixed vs. Decaying $N_L$.} "Decay" indicates a linear reduction of Langevin steps over iterations, adjusted to meet a reduced total budget ($N_T=10$) in the final sampling block. The results show that tapering off exploration in later stages maintains performance.}
  \label{langevin_decaying}
  \centering
  \tiny
  \vspace{0.5em}
  \resizebox{0.7\linewidth}{!}{
  \begin{tabular}{c|c|cccc}
    \toprule
    \textbf{Task} & \textbf{$N_L$} & \textbf{PSNR} ($\uparrow$) & \textbf{SSIM} ($\uparrow$) & \textbf{FID} ($\downarrow$) & \textbf{LPIPS} ($\downarrow$) \\
    \midrule
    \multirow{2}{*}{\shortstack{Inpainting \\ (Box)}}  
    & Fixed ($4$) & 21.89 & \textbf{0.854} & \textbf{37.40} & \textbf{0.101} \\
    & Decay ($4\rightarrow0$) & 21.68 & 0.842 & \textbf{37.00} & 0.108\\
    \midrule
    \multirow{2}{*}{\shortstack{Gaussian \\ Deblurring}}  
    & Fixed ($6$) & 27.83 & 0.744 & \textbf{53.83} & \textbf{0.126} \\
    & Decay ($6\rightarrow1$) & \textbf{27.86} & \textbf{0.750} & 54.13 & 0.129\\
    \midrule
    \multirow{2}{*}{\shortstack{Motion \\ Deblurring}} 
    & Fixed ($6$) & 29.86 & 0.780 & \textbf{35.97} & \textbf{0.065} \\
    & Decay ($6\rightarrow1$) & \textbf{30.06} & \textbf{0.790} & 36.94 & 0.067 \\
    \midrule
    \multirow{2}{*}{\shortstack{Super-Resolution \\ ($\times12$)}}  
    & Fixed ($4$) & \textbf{27.12} & 0.720 & 62.44 & \textbf{0.154} \\
    & Decay ($4\rightarrow0$) & 27.09 & \textbf{0.724} & \textbf{62.38} & 0.156 \\
    \bottomrule
  \end{tabular}
}
\end{table}

\begin{figure}[!h]
    \centering
    \includegraphics[width=0.7\linewidth]{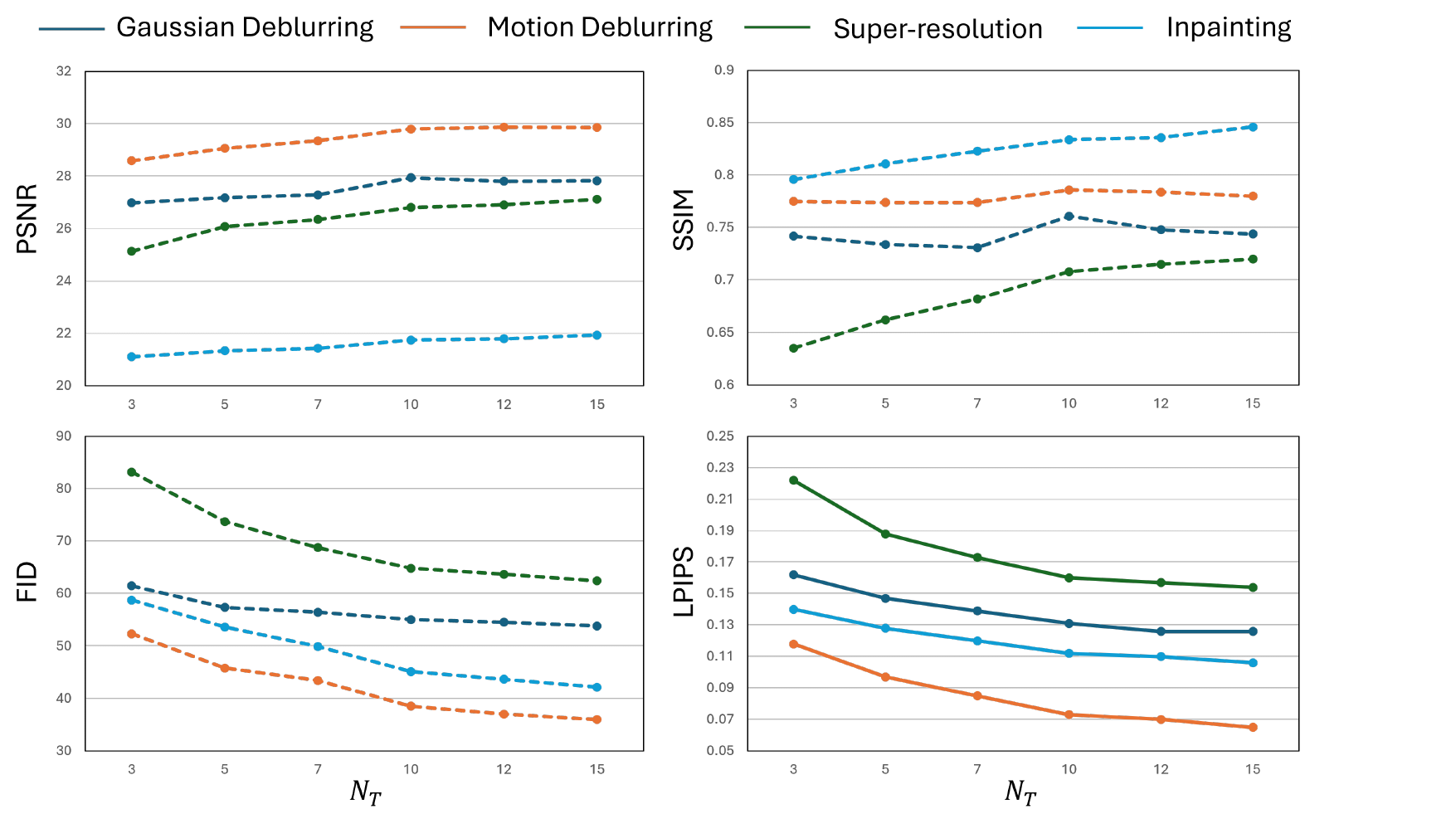}
    \caption{\textbf{Performance vs. Total Steps ($N_T=N_P+N_L$).} Performance gains show diminishing returns beyond $N_T \approx 10$, suggesting that $N_T=15$ serves as a robust operating point for maximum quality, while lower $N_T$ remains viable for efficiency.}
    \label{fig:ablation_N_P}
\end{figure}

\paragraph{Impact of Total Budget ($N_T$).}
Separately, we investigated the trade-off between performance and the total number of Langevin-Proximal steps $N_T=N_L+N_P$.
Fig.~\ref{fig:ablation_N_P} indicates that performance gains show diminishing returns beyond $N_T \approx 10$. While we utilized a fixed $N_T=15$ in our main experiments to prioritize maximum quality, these results suggest that a lower budget (e.g., $N_T=10$) remains a viable choice for time-constrained applications.

\paragraph{Runtime Analysis.}
To quantify practical efficiency, we measured inference time on the inpainting task using an NVIDIA RTX 3090 GPU. We evaluated two VAE backbones: the standard "Large VAE" (SD3) and the distilled "Tiny VAE"~\citep{erbach2025flair}.
Table~\ref{table:runtime_analysis} summarizes the results.
The column "$15 \to 10$" represents the $N_L$ decay strategy.
As shown, the decay strategy reduces inference time compared to the fixed $N_T=15$ setting (e.g., 262.2s $\to$ 232.7s with Large VAE) while preserving quality.
Furthermore, utilizing the Tiny VAE significantly accelerates the process (down to $\approx 30$s), demonstrating that FlowLPS can be substantially accelerated with an efficient decoder.

\begin{table}[h]
  \caption{\textbf{Runtime analysis on inpainting task.} We compare the inference time (seconds per image) across different total budgets ($N_T$). "$15 \to 10$" denotes the dynamic decay strategy where $N_L$ is linearly reduced. "Large VAE" refers to the standard Stable Diffusion VAE, while "Tiny VAE" refers to the distilled efficient decoder used in FLAIR.}
  \label{table:runtime_analysis}
  \tiny
  \centering
  \vspace{0.5em}
  \resizebox{0.65\linewidth}{!}{
  \begin{tabular}{c|ccccccc}
    \toprule
    \textbf{$N_T$} & $3$ & $5$ & $7$ & $10$ & $12$ & $15\rightarrow10$ & $15$ \\
    \midrule
    large VAE & 66.13 & 101.1 & 161.6 & 183.8 & 217.8 & 232.7 & 262.2 \\
    \midrule
    tiny VAE & 15.63 & 18.48 & 21.79 & 24.74 & 27.70 & 29.34 & 31.65 \\
    \bottomrule
  \end{tabular}
}
\end{table}

\begin{figure*}[h]
    \centering
    \includegraphics[width=\linewidth]{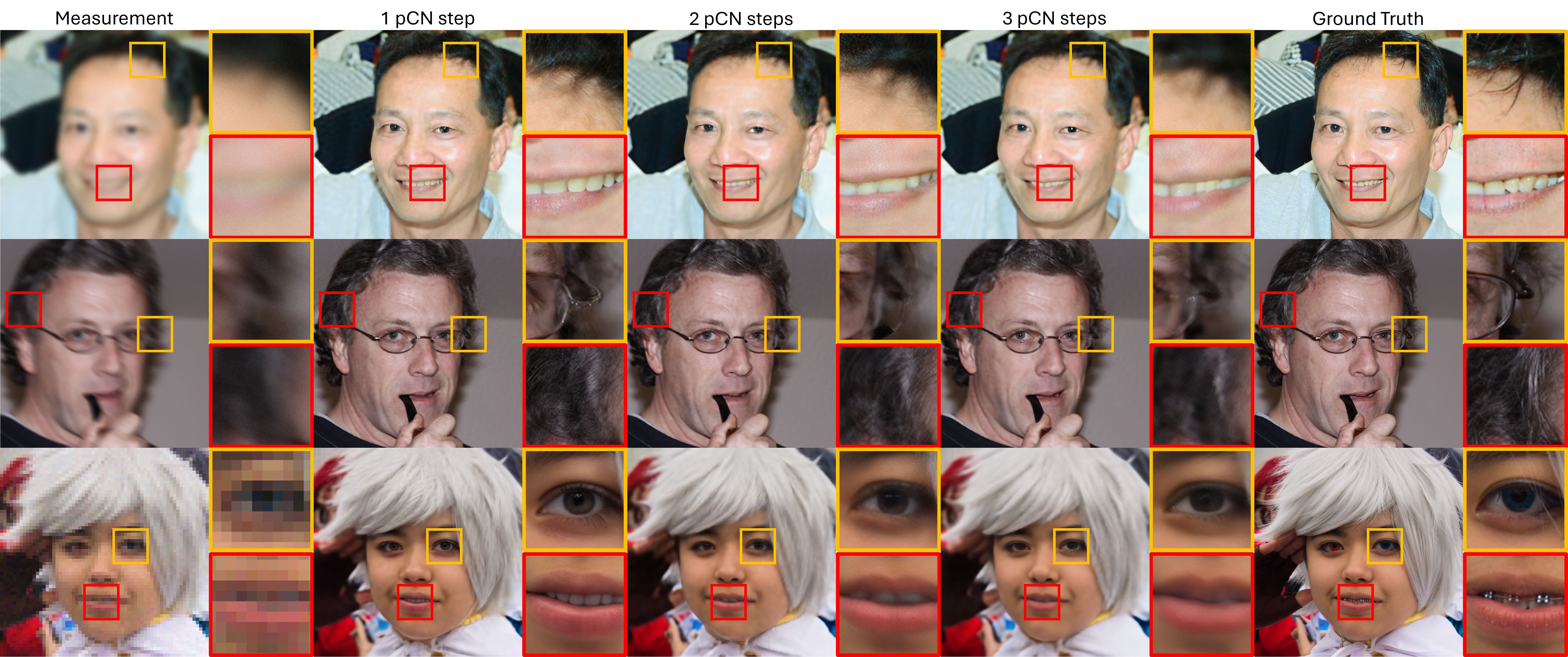}
    \caption{\textbf{Effect of multiple pCN steps.} Increasing pCN steps leads to progressive over-smoothing.}
    \label{fig:number_of_pCN}
\end{figure*}

\subsection{Detailed Analysis of pCN Steps}
\label{num_of_pCN_steps}

\subsubsection{Impact of Number of pCN Re-noising Steps.}
For completeness, we recall the Gaussian-preserving property underlying the pCN-style refresh.
If the target distribution is the standard Gaussian $\mathcal{N}(\mathbf{0},\mathbf{I})$, the corresponding pCN proposal has acceptance probability:
\begin{align}
    &\min \Big\{1, \frac{p_{\text{tar}}(\hat\vz_{\text{new}})q_{\text{prop}}(\hat\vz_{\text{old}}|\hat\vz_{\text{new}})}{p_{\text{tar}}(\hat\vz_{\text{old}})q_{\text{prop}}(\hat\vz_{\text{new}}|\hat\vz_{\text{old}})}\Big\}\\
    =&\min \Big\{1, \frac{\mathcal{N}(\hat\vz_{\text{new}};0,I)\mathcal{N}(\hat\vz_{\text{old}}; \rho_t \hat\vz_{\text{new}}, (1-\rho_t^2)I)}{\mathcal{N}(\hat\vz_{\text{old}};0,I)\mathcal{N}(\hat\vz_{\text{new}};\rho_t \hat\vz_{\text{old}},(1-\rho_t^2)I)}\Big\} \nonumber\\
    =&\min\{1,1\} = 1. \nonumber
\end{align}
Thus, in this idealized Gaussian case, every pCN proposal is accepted.
This idealized property motivates the variance-preserving form of the refresh, although FlowLPS initializes it from the model-predicted noise in practice.

Table~\ref{tab:renoising_steps} compares performance across varying numbers of pCN steps.
We observe that increasing the number of pCN steps slightly improves PSNR but degrades perceptual metrics.
As visualized in Fig.~\ref{fig:number_of_pCN}, multiple pCN steps dilute the structural information embedded in the predicted noise $\hat\vz_{1|t}^{(0)}$, leading to progressive over-smoothing.

Hypothesizing that stronger stochastic refresh is beneficial primarily in early stages, we further experiment with a linear decay schedule for pCN steps ($4 \rightarrow 1$).
As shown in Table~\ref{tab:renoising_steps}, this decay strategy improves FID in generation-heavy tasks such as box inpainting and super-resolution.
However, this perceptual improvement comes at the cost of pixel-level fidelity, and in deblurring tasks the decay strategy offers only marginal benefits over the single-step baseline.
Therefore, considering computational efficiency, consistent performance across tasks, and implementation simplicity, we use a single pCN step as the default choice.

\begin{table}[h]
  \caption{\textbf{Ablation on the number of pCN-style re-noising steps (FFHQ 0.2k).} "Decay ($4\rightarrow 1$)" indicates a linear reduction of pCN steps from 4 to 1 over the sampling process. While the decay strategy offers the best FID for inpainting and super-resolution, the single-step approach provides the best overall balance.}
  \vspace{0.5em}
  \label{tab:renoising_steps}
  \centering
  \tiny
  \resizebox{0.7\linewidth}{!}{
  \begin{tabular}{c|c|cccc}
    \toprule
    \textbf{Task} & \textbf{\# pCN} & \textbf{PSNR} ($\uparrow$) & \textbf{SSIM} ($\uparrow$) & \textbf{FID} ($\downarrow$) & \textbf{LPIPS} ($\downarrow$) \\
    \midrule
    \multirow{4}{*}{\shortstack{Inpainting \\ (Box)}} 
    & 1 & 21.94 & \textbf{0.846} & 42.15 & \textbf{0.106}\\
    & 2 & 21.67 & 0.842 & 42.55 & 0.110\\
    & 3 & \textbf{21.98} & \textbf{0.846} & 43.87 & 0.108\\
    & Decay ($4\rightarrow 1$) & 21.46 & 0.843 & \textbf{40.34} & 0.110\\
    \midrule
    \multirow{4}{*}{\shortstack{Gaussian \\ Deblurring}} 
    & 1 & 27.83 & 0.744 & \textbf{53.83} & \textbf{0.126}\\
    & 2 & 27.97 & 0.750 & 54.62 & 0.129\\
    & 3 & \textbf{28.01} & \textbf{0.752} & 56.36 & 0.129\\
    & Decay ($4\rightarrow 1$) & 27.75 & 0.745 & 54.32 & 0.128\\
    \midrule
    \multirow{4}{*}{\shortstack{Motion \\ Deblurring}} 
    & 1 & 29.86 & 0.780 & \textbf{35.97} & \textbf{0.065}\\
    & 2 & 30.10 & 0.789 & 37.09 & 0.067\\
    & 3 & \textbf{30.13} & \textbf{0.790} & 37.60 & 0.067\\
    & Decay ($4\rightarrow 1$) & 30.02 & 0.786 & 36.90 & 0.067\\
    \midrule
    \multirow{4}{*}{\shortstack{Super-Resolution \\($\times12$)}} 
    & 1 & 27.12 & 0.720 & 62.44 & \textbf{0.154}\\
    & 2 & \textbf{27.14} & \textbf{0.725} & 65.04 & 0.158\\
    & 3 & \textbf{27.14} & \textbf{0.725} & 65.46 & 0.161\\
    & Decay ($4\rightarrow 1$) & 26.92 & 0.715 & \textbf{62.03} & 0.156\\
    \bottomrule
  \end{tabular}
}
\end{table}

\subsubsection{Qualitative Analysis of $\rho_t$ Schedules}
\begin{figure}[h]
    \centering
    \includegraphics[width=0.6\linewidth]{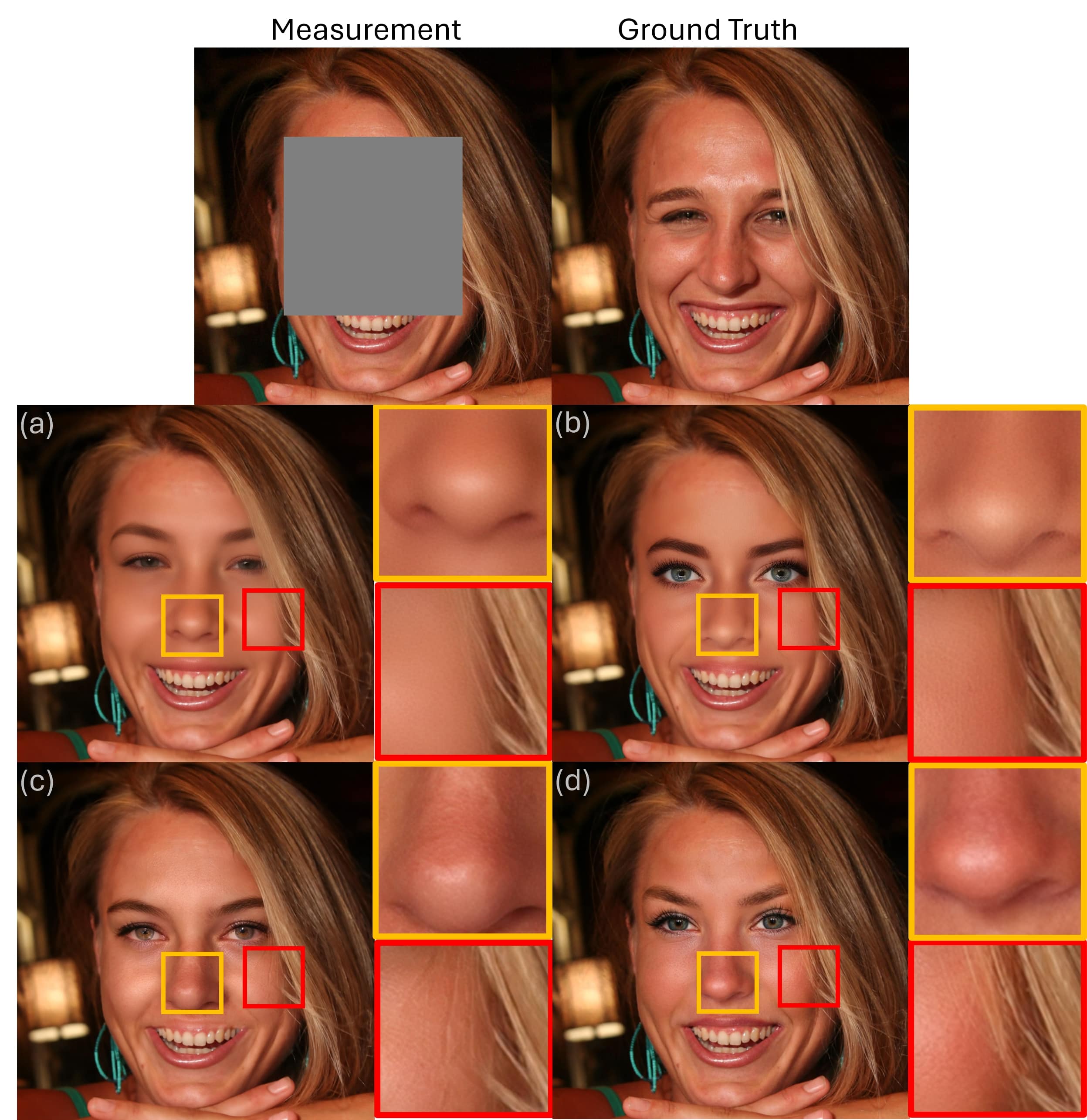}
    \caption{\textbf{Qualitative comparison of $\rho_t$.} (a) $\rho_t=0$, (b) $\rho_t=0.5$, (c) $\rho_t=1-t$, (d) $\rho_t=\sqrt{1-t}$. Our adaptive schedule (d) best preserves high-frequency details and semantic continuity (e.g., hair strands) compared to other schedules.}
    \label{fig:different_rho_t}
\end{figure}

Complementing the quantitative results in Section~\ref{sec:rho_t strategy}, we provide visual comparisons of different $\rho_t$ schedules in Fig.~\ref{fig:different_rho_t}.
As discussed, the adaptive schedule $\rho_t=\sqrt{1-t}$ reconstructs continuous semantic details, such as hair strands across the mask, whereas other schedules often fail to maintain such structural continuity or produce artifacts.
This visual evidence reinforces the advantage of the adaptive mixing schedule shown in Table~\ref{tab:different_rho_t}.

\subsection{Component-wise Effect of Langevin Updates and pCN-style Re-Noising}
\label{sec:component_ablation}

We performed a controlled $2\times 2$ ablation over Langevin updates and pCN-style re-noising, while keeping the proximal stage fixed on the inpainting task.
Table~\ref{tab:component_ablation} shows that both components help and that their combination gives the best perceptual quality.
This suggests that the gain does not come from a single stochastic component: Langevin updates provide posterior-oriented initialization, while pCN-style re-noising supplies complementary stochastic refreshment.
\begin{table}[h]
    \centering
    \caption{\textbf{Component-wise ablation of Langevin updates and pCN-style re-noising on box inpainting.}
Disabling Langevin sets $N_L=0$. Disabling controlled pCN-style re-noising uses independent Gaussian re-noising $(\rho_t=0)$.
Both components improve FID, and their combination yields the strongest perceptual quality under the fixed update budget.}
    \label{tab:component_ablation}
    \vspace{0.5em}
    \scriptsize
    \begin{tabular}{cc|cc}
    \toprule
    pCN re-noising & Langevin & PSNR $\uparrow$ & FID $\downarrow$ \\
    \midrule
    \xmark & \xmark & 22.62 & 45.37 \\
    \cmark & \xmark & 21.88 & 28.81 \\
    \xmark & \cmark & \textbf{22.74} & 30.95 \\
    \cmark & \cmark & 21.68 & \textbf{21.90} \\
    \bottomrule
    \end{tabular}
\end{table}

\section{Additional Qualitative Results}
\subsection{Visual Analysis of Generative Bias in FLAIR}
\label{sec:flair_analysis}
\begin{figure}[h]
    \centering
    \includegraphics[width=0.75\linewidth]{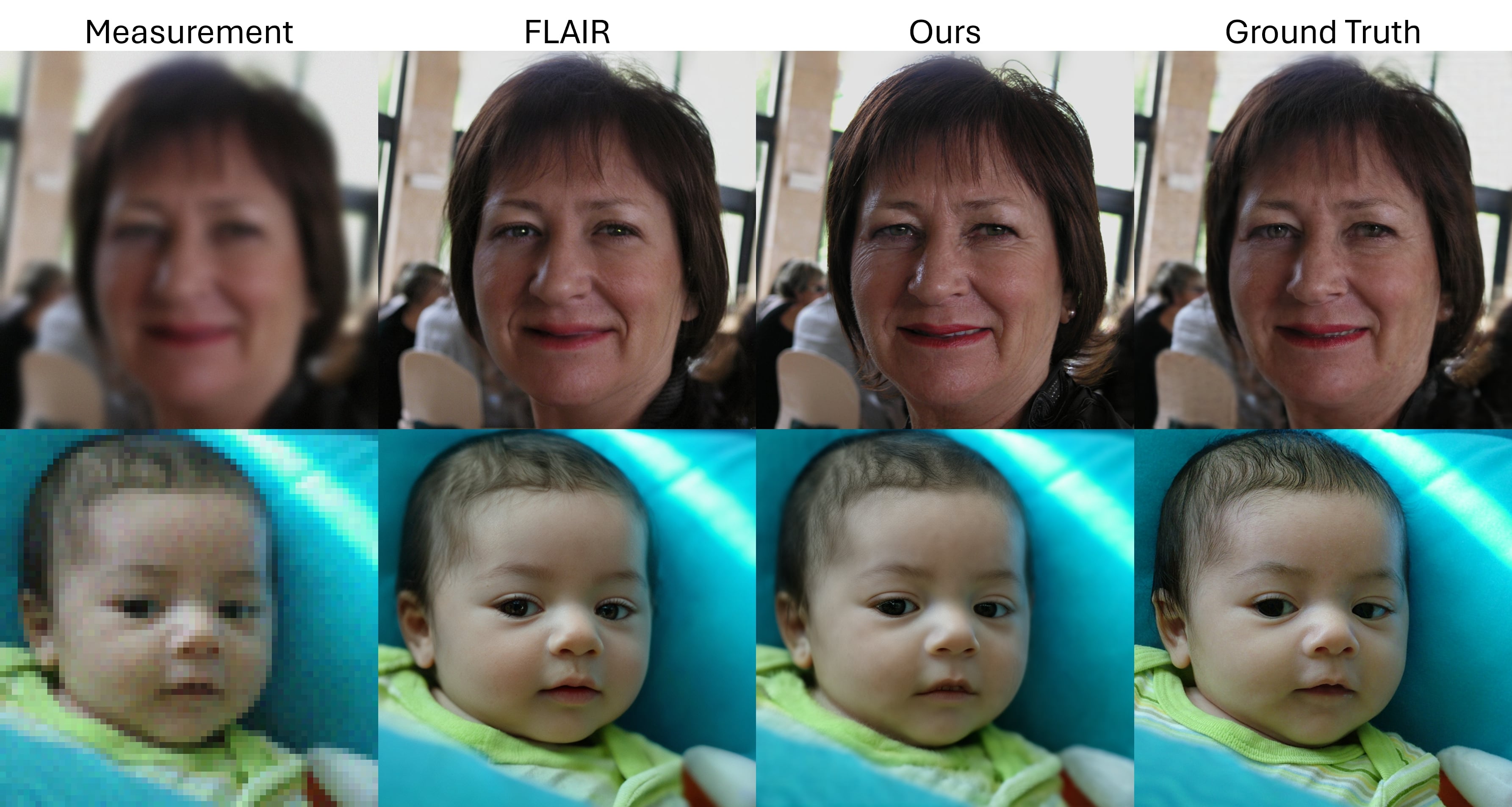}
    \caption{\textbf{Visual evidence of generative bias in FLAIR.} Top: Gaussian Deblurring, Bottom: Super-Resolution. 
    While FLAIR generates sharp images, it ignores the measurement's underlying structure in favor of its generative prior. 
    Note how FLAIR alters facial features (e.g., eye shape), prioritizing the generation of "plausible" faces over the recovery of the true identity. 
    In contrast, FlowLPS better balances prior plausibility with measurement-consistent semantic structure.}
    \label{fig:compare_flair}
\end{figure}

As discussed in the main paper (Sec.~\ref{sec:main_results}), FLAIR can produce sharp but less measurement-faithful reconstructions.
Fig.~\ref{fig:compare_flair} presents a detailed visual comparison.
While FLAIR generates sharp images, it can alter semantic attributes such as facial expressions or specific features (e.g., eye shape, wrinkles), suggesting a stronger reliance on the generative prior than on the measurement.
In contrast, FlowLPS better balances prior plausibility with measurement-consistent semantic structure.

\subsection{Sample Diversity}
\label{sec:sample_diversity}

While FlowLPS includes a proximal refinement step, it remains different from deterministic refinement-dominant solvers.
Its stochasticity comes from two sources: Langevin updates during posterior-oriented initialization and pCN-style re-noising along the reverse trajectory.
These components allow different random seeds to produce distinct posterior-oriented initializations, after which the proximal step refines each trajectory toward a nearby measurement-consistent reconstruction.
Thus, repeated runs can yield diverse valid solutions rather than collapsing to a single deterministic output.

To quantitatively evaluate diversity, we follow the RLSD protocol.
For each measurement, we generate 32 samples, retain only samples whose measurement residual is below a fixed validity threshold, and compute the average pairwise cosine similarity of DINO features among the valid samples; lower similarity indicates higher diversity.
We also report the valid-sample ratio, since diversity among measurement-inconsistent samples can be misleading.

\begin{table}[h]
  \caption{\textbf{Diversity analysis on the inpainting task.} 
  We report pairwise DINO cosine similarity among valid samples and the valid-sample ratio. Lower cosine similarity indicates higher diversity.}
  \label{table:diversity_analysis}
  \tiny
  \centering
  \vspace{0.5em}
  \resizebox{0.7\linewidth}{!}{
  \begin{tabular}{c|ccccc}
    \toprule
    Methods & DAPS (SD3) & FlowDPS & FlowChef & Flower & FlowLPS (Ours) \\
    \midrule
    Pairwise DINO cosine $\downarrow$ & 0.9681 & 0.9215 & 0.8341 & 0.9823 & 0.9714 \\
    \midrule
    Valid-sample ratio $\uparrow$ & 1.0 & 0.47 & 0.12 & 1.0 & 1.0 \\
    \bottomrule
  \end{tabular}
}
\end{table}

We omit RLSD because it rarely produced valid samples under the same threshold, making its diversity score unreliable.
FlowDPS and FlowChef achieve lower pairwise similarity, but their valid-sample ratios are substantially lower, indicating that much of their apparent diversity comes from measurement-inconsistent reconstructions.
By contrast, FlowLPS maintains a 100\% valid-sample ratio while retaining competitive diversity.
Among methods with a 100\% valid-sample ratio, FlowLPS is much closer to DAPS than to the refinement-dominant baseline Flower, suggesting that proximal refinement does not reduce FlowLPS to a deterministic refinement-only solver.

\begin{figure}
    \centering
    \includegraphics[width=\linewidth]{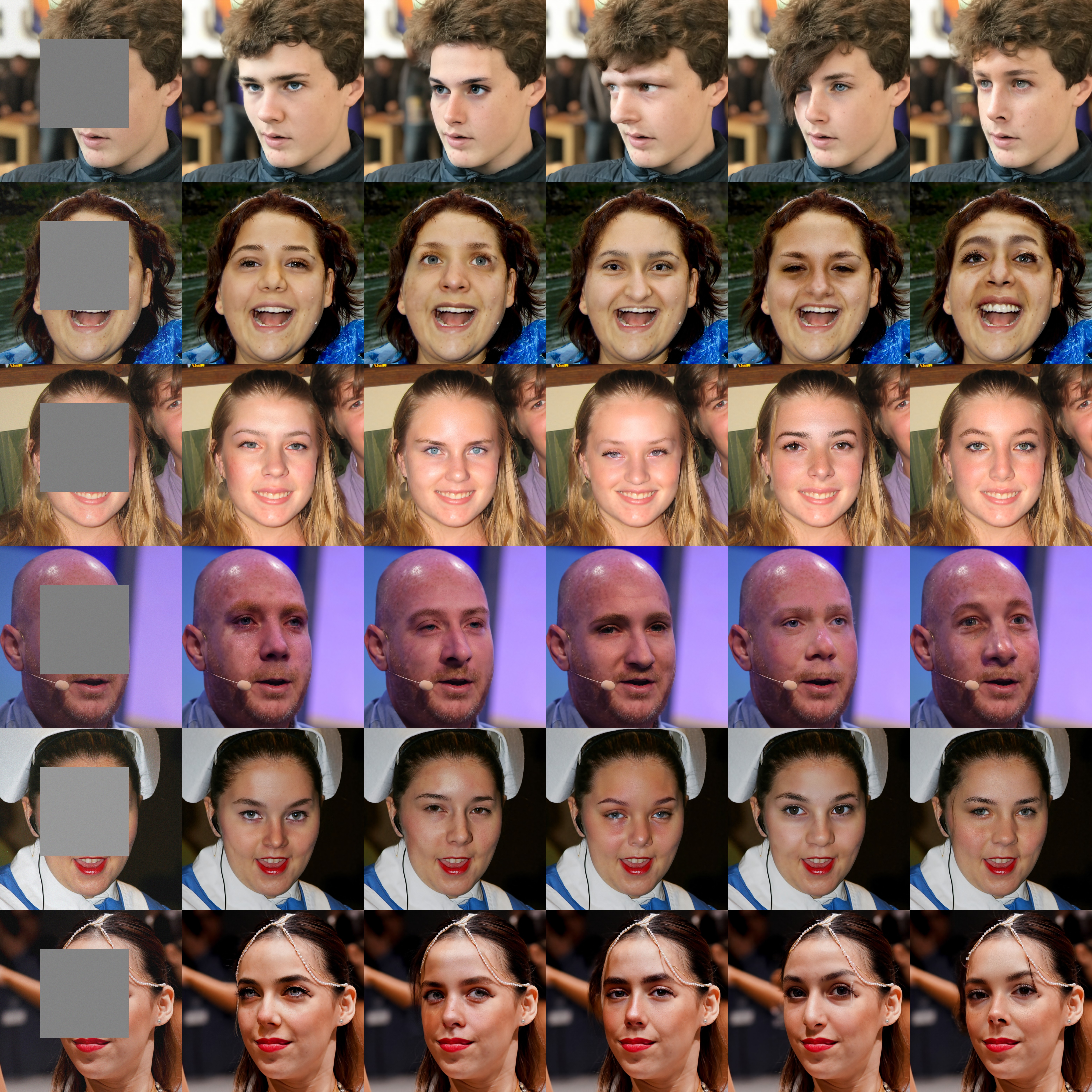}
    \caption{\textbf{Diversity of solutions in box inpainting.}
Given the same masked input (leftmost column), FlowLPS generates distinct yet plausible measurement-consistent reconstructions.
The results suggest that the Langevin-Proximal design can produce multiple valid completions rather than collapsing to a single deterministic reconstruction.}
    \label{fig:diversity}
\end{figure}

Figure~\ref{fig:diversity} demonstrates this behavior on box inpainting.
Given the same masked input, FlowLPS generates multiple valid reconstructions with meaningful variations in high-frequency details while remaining consistent with the unmasked regions.
This suggests that FlowLPS preserves stochastic solution diversity while maintaining measurement consistency.

\subsection{Full Reconstruction Results}
\label{additional_results}

Additional qualitative comparisons on FFHQ and DIV2K across all tasks are provided in the supplementary material.
These results complement the representative visual comparisons in the main paper and further illustrate that FlowLPS preserves sharper textures and natural local details while maintaining measurement consistency.

\begin{figure*}
    \centering
    \includegraphics[width=\linewidth]{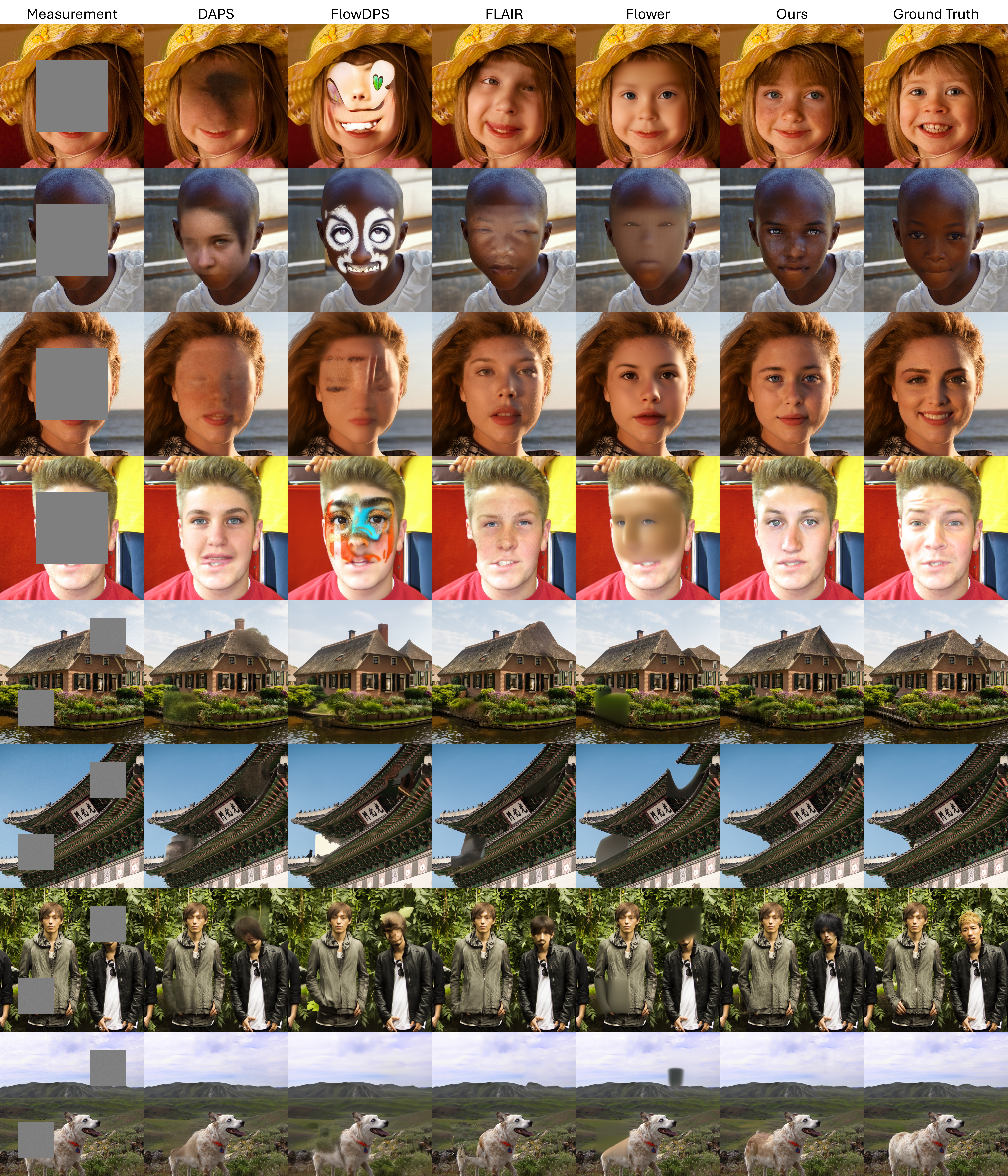}
    \caption{Box Inpainting results on FFHQ and DIV2K.}
    \label{fig:box_inpainting}
\end{figure*}
\begin{figure*}
    \centering
    \includegraphics[width=\linewidth]{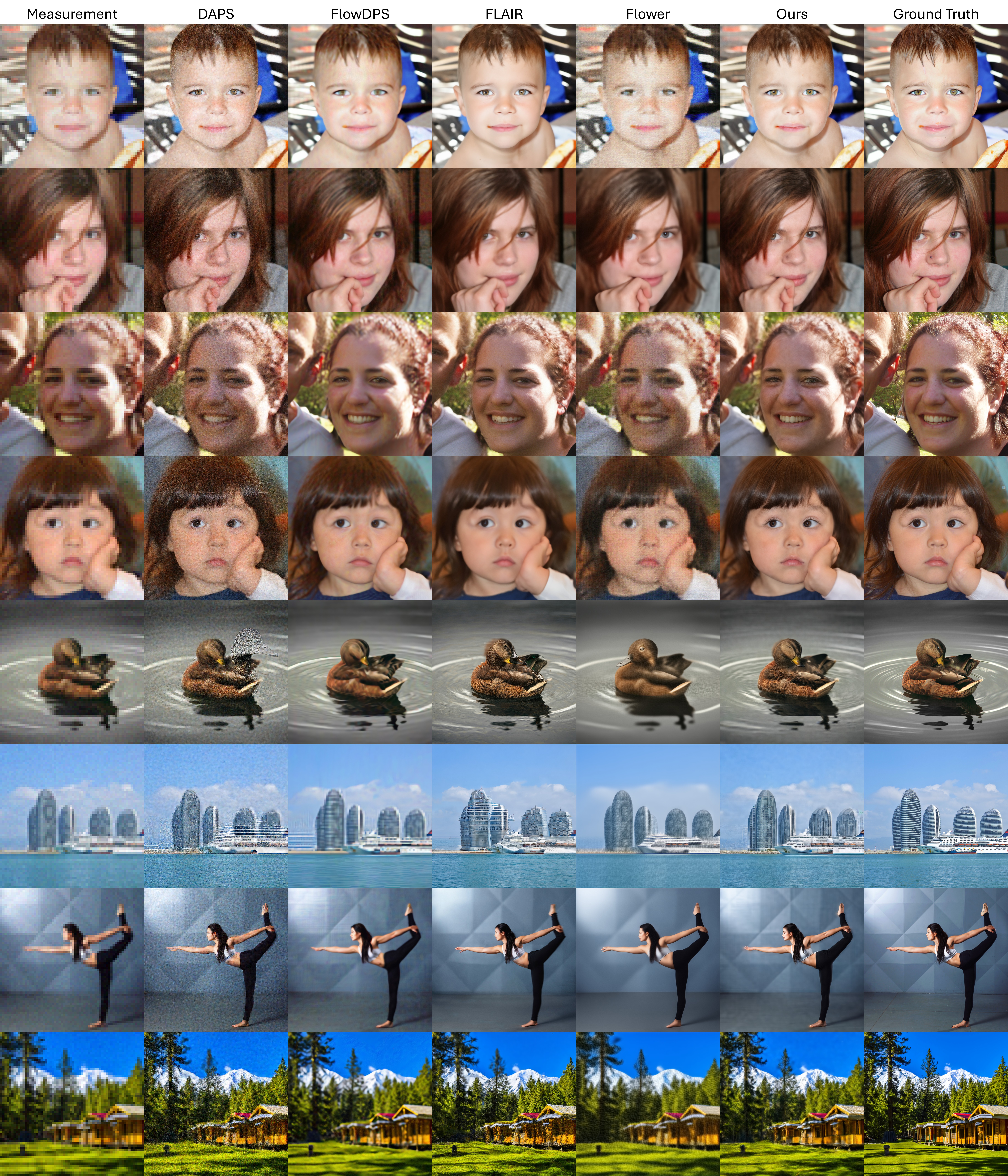}
    \caption{Super-Resolution($\times 12$) results on FFHQ and DIV2K.}
    \label{fig:sr}
\end{figure*}
\begin{figure*}
    \centering
    \includegraphics[width=\linewidth]{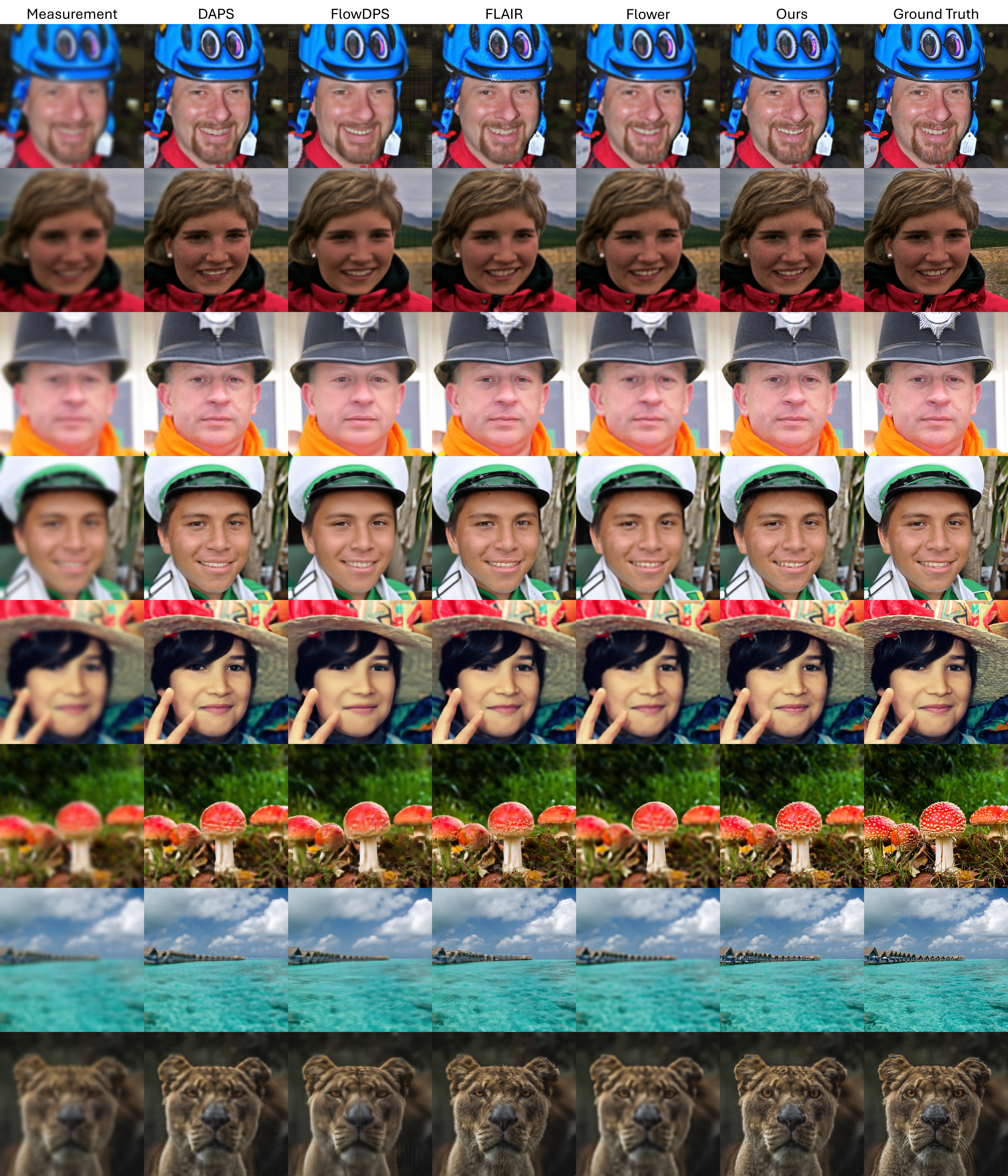}
    \caption{Gaussian deblurring results on FFHQ and DIV2K.}
    \label{fig:gaussian_deblur}
\end{figure*}
\begin{figure*}
    \centering
    \includegraphics[width=\linewidth]{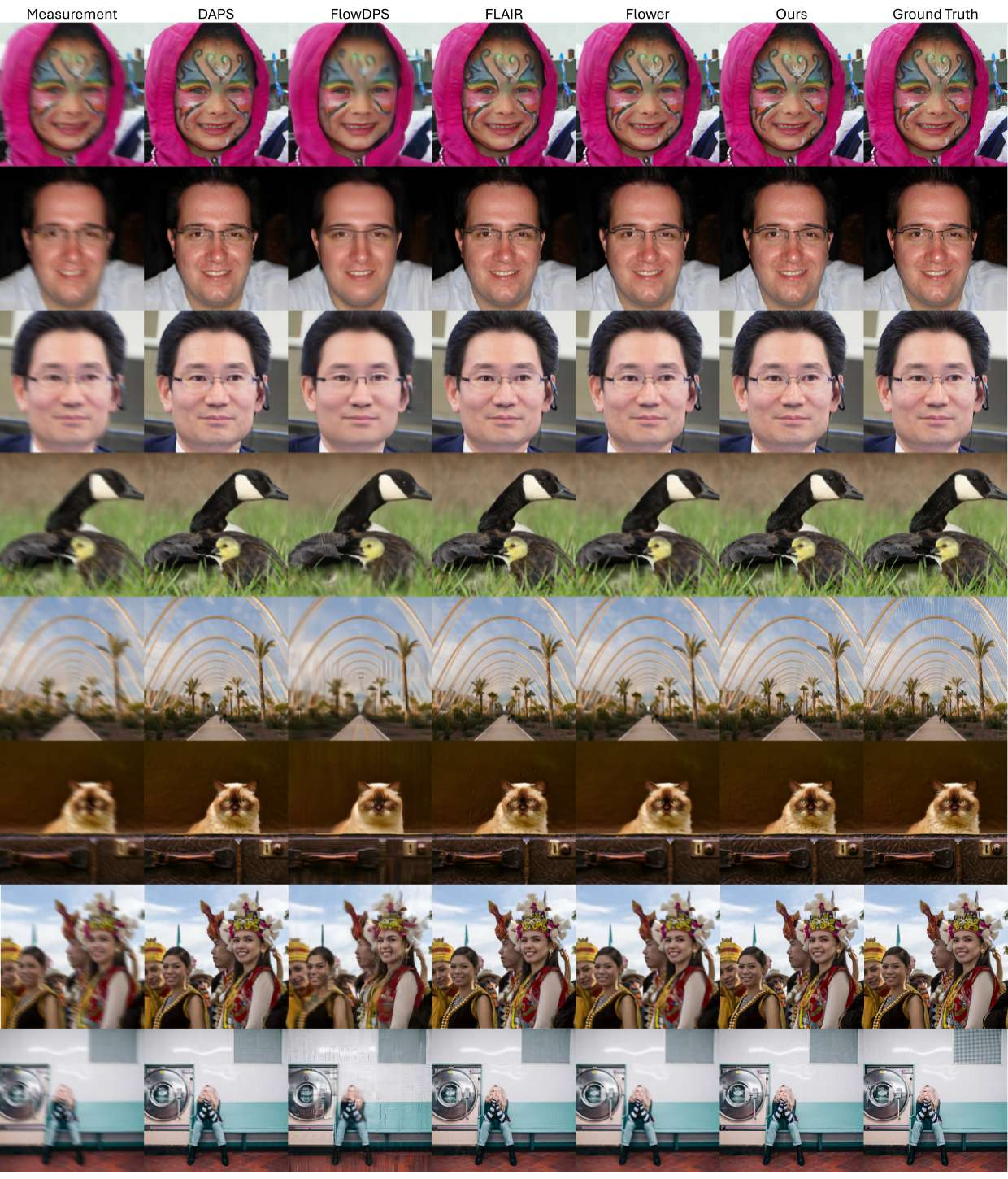}
    \vspace{-2cm}
    \caption{Motion deblurring results on FFHQ and DIV2K.}
    \label{fig:motion_deblur}
\end{figure*}
\begin{figure*}
    \centering
    \includegraphics[width=\linewidth]{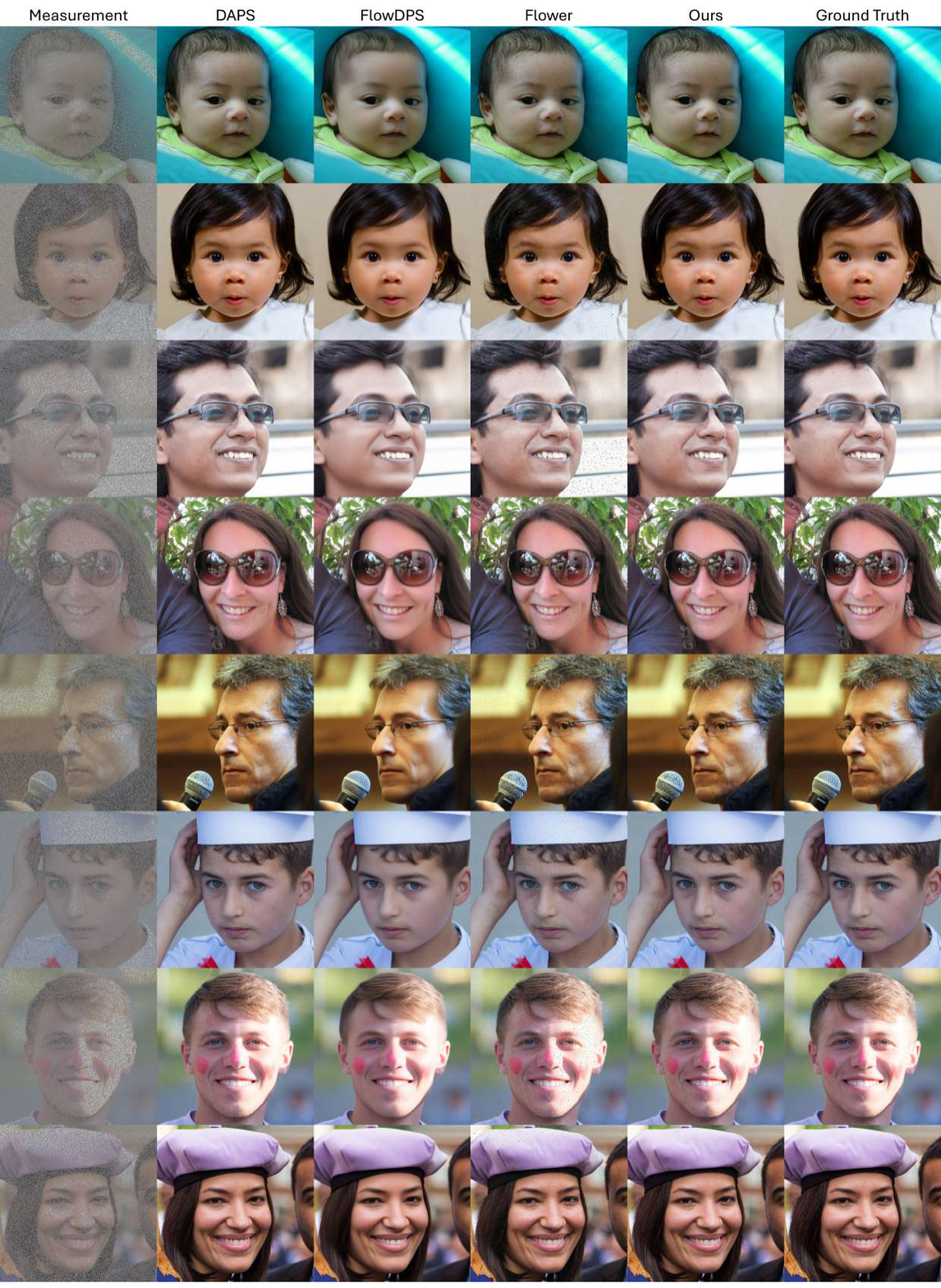}
    \vspace{-0.7cm}
    \caption{Random Inpainting results on FFHQ and DIV2K.}
    \label{fig:random_inpainting}
\end{figure*}


\end{document}